\documentclass[10pt,twocolumn,letterpaper]{article}

\usepackage{wacv}              
\usepackage[accsupp]{axessibility} 
\usepackage{graphicx}
\usepackage{amssymb}
\usepackage{booktabs}

\usepackage{algorithmic}
\usepackage{float}
\usepackage{nameref}
\usepackage[flushleft]{threeparttable}
\usepackage{color}
\usepackage{colortbl}
\usepackage{multirow}
\usepackage{array}
\usepackage[table,dvipsnames]{xcolor}

\usepackage[toc,page,header]{appendix}
\usepackage{minitoc}

\usepackage{booktabs}

\usepackage{amsmath,amsbsy}
\DeclareUnicodeCharacter{03B8}{$\theta$}

\usepackage[pagebackref,breaklinks,colorlinks]{hyperref}

\usepackage[linesnumbered,ruled,vlined]{algorithm2e}

\SetKwInput{KwInput}{Input}
\SetKwInput{KwOutput}{Output}

\usepackage[capitalize]{cleveref}
\crefname{section}{Sec.}{Secs.}
\Crefname{section}{Section}{Sections}
\Crefname{table}{Table}{Tables}
\crefname{table}{Tab.}{Tabs.}

\def\wacvPaperID{571} 
\def\confName{WACV}
\def\confYear{2025}

\begin{document}
	
	
	
	\makeatletter
	\DeclareRobustCommand\onedot{\futurelet\@let@token\@onedot}
	\def\@onedot{\ifx\@let@token.\else.\null\fi\xspace}
	
	\def\eg{\emph{e.g}\onedot} \def\Eg{\emph{E.g}\onedot}
	\def\ie{\emph{i.e}\onedot} \def\Ie{\emph{I.e}\onedot}
	\def\cf{\emph{c.f}\onedot} \def\Cf{\emph{C.f}\onedot}
	\def\etc{\emph{etc}\onedot} \def\vs{\emph{vs}\onedot}
	\def\wrt{w.r.t\onedot} \def\dof{d.o.f\onedot}
	\def\etal{\emph{et al}\onedot}
	\makeatother
	
	\title{Multi-Surrogate-Teacher Assistance for Representation Alignment in Fingerprint-based Indoor Localization}
	
	\author{Son Minh Nguyen\textsuperscript{1}, Linh Duy Tran\textsuperscript{2}, Duc Viet Le\textsuperscript{1}, Paul J.M Havinga\textsuperscript{1}\\
		\textsuperscript{1}Department of Computer Science, University of Twente\\
		\textsuperscript{2}Viettel AI, Viettel Group\\
		{\tt\small \{m.s.nguyen, v.d.le, p.j.m.havinga\}@utwente.nl},
		{\tt\small linhtd15@viettel.com.vn}
}
\maketitle

\begin{abstract}
	Despite remarkable progress in knowledge transfer across visual and textual domains, extending these achievements to indoor localization, particularly for learning transferable representations among Received Signal Strength (RSS) fingerprint datasets, remains a challenge.
	This is due to inherent discrepancies among these RSS datasets, largely including variations in building structure, the input number and disposition of WiFi anchors\footnote{An anchor describes a radio-emitting source. (\eg, WiFi access points, Bluetooth beacons.)}. 
	Accordingly, specialized networks, which were deprived of the ability to discern transferable representations, readily incorporate environment-sensitive clues into the learning process, hence limiting their potential when applied to specific RSS datasets.
	In this work, we propose a plug-and-play (PnP) framework of knowledge transfer, facilitating the exploitation of transferable representations for specialized networks directly on target RSS datasets through two main phases.
	Initially, we design an \textit{Expert Training} phase, which features multiple surrogate generative teachers, all serving as a global adapter that homogenizes the input disparities among independent source RSS datasets while preserving their unique characteristics. In a subsequent \textit{Expert Distilling} phase, we continue introducing a triplet of underlying constraints that requires minimizing the differences in essential knowledge between the specialized network and surrogate teachers through refining its representation learning on the target dataset. This process implicitly fosters a representational alignment in such a way that is less sensitive to specific environmental dynamics.
	Extensive experiments conducted on three benchmark WiFi RSS fingerprint datasets underscore the effectiveness of the framework that significantly exerts the full potential of specialized networks in localization\footnote{Our code is available at \url{https://github.com/Minh-Son-Nguyen/RSS_TL}.}.
\end{abstract}

\vspace{-10pt}
\section{Introduction}
\label{sec:intro}
\vspace{-5pt}
The proliferation of WiFi, and Bluetooth infrastructures has propelled Received-Signal-Strength (RSS) fingerprint-based indoor localization into the spotlight of both academic and industrial communities. This surge of interest is immensely driven by the rising demand for location-based services, such as asset tracking, wayfinding, and patient monitoring. Lately, the roaring success of deep learning has invigorated this field further by dint of its relevance to mainstream applications in Computer Vision, and Natural Language Processing. In this context, fingerprinting approaches initially require a collection phase to build an RSS fingerprint database that is normally constituted by sequences of recorded RSS fingerprints together with associated locations for a given area. Subsequently, various machine learning algorithms are utilized to learn a matching function that expresses the correlation between RSS fingerprints and their associated locations. In the inference phase, individuals' locations are determined from query RSS fingerprint observations using the learned matching function.

Over the past few years, fingerprint-based indoor localization has undergone a complete transformation from deterministic KNN-based methods\cite{bahl2000radar,gu2016reducing,xie2016improved} to sophisticated neural architectures, \eg, RNNs\cite{jang2017geomagnetic}, CNNs\cite{song2019novel, shao2018indoor, jang2018indoor}, and more recently Transformers\cite{nguyen2023learning}. Although significant efforts have gone into utilizing these architectures to extract robust features for enhanced localization accuracy, knowledge transfer paradigms that could maximize the potential of these specialized networks better, as yet, only exist in embryonic form.

In particular, based on the radio multipath propagation, RSS fingerprint datasets are deliberately characterized by distinct structural setups, \eg, the input number and arrangement of WiFi anchors, inner infrastructures, and static layouts of surrounding objects. However, this radio feature, which contributes to the specificity of the datasets, is also deemed a double-edged sword. The susceptibility of RSS fingerprints to environmental changes\cite{li2021train} (\eg, movable subjects, temperature, and humidity), as a result of the radio multipath effect, greatly hinders progress in this domain. In contrast to modalities like images, where context remains visually consistent under different circumstances, RSS fingerprints linked to specific locations might be valid within a particular timeframe and confined to a certain area. Such unique characteristics induce nonconformity in the input size and expressiveness between the RSS datasets. Accordingly, existing neural models currently used in this domain are highly specialized to their respective datasets, and the potential knowledge they acquired from other RSS fingerprint datasets remains largely unexplored and isolated.

As a result, these discrete and location-specific (\ie \textit{independent}) RSS datasets render classical transfer-learning frameworks, whether implemented \textit{sequentially} or \textit{collectively}, impractical to be utilized for other RSS datasets. 
Our proposed approach operates on the assumption that a specialized model can \textit{partially} and \textit{implicitly} share its underlying ability to learn representations from various source RSS datasets, offering advantages in enhancing localization accuracy on the target dataset.
The core idea is grounded by the fact that each RSS dataset is influenced by unique environmental dynamics at different levels. By maintaining alignment with a broad spectrum of learning abilities acquired from various source datasets simultaneously, the model is encouraged to prioritize the learning of \textit{transferable} representations that should ideally be \textit{independent of} or, at the very least, \textit{less sensitive} to specific environmental dynamics, thus leading to improved localization accuracy. For this purpose, we propose a plug-and-play (PnP) framework, the first of its kind, promoting the learning of transferable representations in two key phases by achieving a comprehensive representation alignment between the RSS target and multiple source RSS datasets.

First, we introduce an \textit{Expert Training} phase as a global adapter designed to work around uncompromising discrepancies across distinct source datasets. Rather than directly modeling the raw RSS data, which was notorious for its variability over time, we attend to the essence of representations learned by specialized networks, which guides us to less noisy and more concise information. This phase involves the deployment of multiple generative models, each associated with one specific source dataset, functioning as surrogate-teacher networks modeling representations established by the specialized network on the corresponding source datasets. By harnessing individual surrogate generative teachers for each source dataset, we can achieve a dynamic, yet homogeneous dimensional space of all pre-established representations from various source datasets where the unique characteristics inherent in each of the source datasets are retained.

Second, we design an \textit{Expert Distilling} phase to establish an alignment of representations learned on the target dataset simultaneously with those modeled from the source datasets. To enhance the alignment, where irrelevant information is filtered out, and only essential knowledge is transferred, we define a triplet based on basic information theories 
: (i) Angular Similarity $J_{Sim}$ to guarantee heightened correlation between specialized representations learned on the target dataset and those modeled from source datasets, (ii) Cross-Mutual Information $J_{MI}$ to capture the necessary complementary information between the specialized representations and their 
modeled counterparts, and finally (iii) Functional Information $J_{FI}$, a Lipschitz measure accounting for the functional relationship between the specialized network and surrogate-teacher networks.
Since the triplet implicitly adjusts the alignment on essential knowledge with some degree of adaptability to different datasets, the learning of transferable representations is less affected by the relevance and quality of involved source datasets. 

In summary, our main contributions are as follows:
\vspace{-3pt}
\begin{itemize}
	\item To our best knowledge, this framework is the first successful attempt that accomplishes knowledge transfer across RSS fingerprint datasets without necessitating significant alterations to existing architectures or additional target data for performance gains. The primary intention is to create a representational alignment between the target RSS dataset and source datasets, through which the learning of environment-insensitive (\ie \textit{transferable}) representations is greatly stimulated. 
	\item The modeling and extraction of essential knowledge from multiple independent source datasets, followed by its transfer to the specialized network, is orchestrated in a novel scheme, where knowledge transfer is merely achieved under the tutelage of surrogate-teacher networks without the need to access the source data. This PnP scheme also opens up opportunities for various communities with data privacy concerns, such as those in biometrics and medical imaging, to attain optimal performance.
	
	\item Extensive experiments are rigorously conducted on SoTA specialized networks over three benchmark RSS datasets to confirm our efficacy, considering the quality and relevance of the source RSS datasets to the target.
\end{itemize}
\vspace{-15pt}
\section{Related Work}
\vspace{-4pt}
\textbf{Fingerprint-based Indoor Localization.} 
In this category, methods aim to learn matching functions describing the relationship between RSS fingerprints and associated locations. In the early stages of development, $k$NN-based methods \cite{bahl2000radar,gu2016reducing,xie2016improved} driven by various distance metrics (\eg, Euclidean, Manhattan, and Cosine metrics) were widely used to determine locations from $k$ nearest RSS fingerprints to the query fingerprint observations. After that, Gibr{\'a}n \textit{et al.}\cite{felix2016fingerprinting} employed DNN-based variants to enhance estimation accuracy. 
Recent approaches\cite{sinha2021completely,song2019novel,jang2018indoor} transformed one-dimensional RSS fingerprint sequences into two-dimensional RSS images to fully utilize architectural inductive biases from CNNs for feature exploitation. Not long ago, Nguyen \textit{et al.}\cite{nguyen2023learning} achieved impressive performance with Transformer-based networks capable of capturing distinct feature representations embedded in RSS fingerprint sequences for specific locations.

\noindent\textbf{Knowledge Distillation.} 
The theoretical underpinnings of distillation have a long history dating back to the work\cite{craven1995extracting,liang2008structure,vapnik2015learning}. However, this research was only actively revisited and on the upswing right after the work introduced by Hinton \textit{et al.}\cite{hinton2015distilling} in the context of model compression. This seminal knowledge distillation initially transfers rich knowledge from a strong teacher model to a target student model based on its soft probability distribution of the target task, which encourages the student model to learn how to mimic the task-specific predictability of the teacher model. This expectantly brings the student's performance closer to the teacher model with no parameter footprint added. Beyond usual soft probabilistic labels, later methods discovered new types of knowledge, \eg, intermediate feature maps\cite{romero2014fitnets}, attention maps\cite{zagoruyko2016paying}, and activation boundaries\cite{heo2019knowledge}. In the most current and basic form in fingerprint-based indoor localization, it was presented by Mazlan \textit{et al.}\cite{mazlan2022fast} to preserve the localizing performance when embedded on computation-constrained devices.

\noindent\textbf{Transfer-Learning.}  Through the years, this paradigm has indeed proved efficient and effective; in particular, outstanding work\cite{kolesnikov2020big,dosovitskiy2020image} showed strong performance when fine-tuned on downstream vision tasks using pre-trained models on JFT-300M dataset\cite{sun2017revisiting}. In contrast, transfer learning for RSS fingerprint-based localization is largely unexplored. Most transfer learning methods work out under the assumption that the source datasets and the target dataset should establish feature spaces in close proximity, but this is not the case for such RSS fingerprint datasets due to their unique characteristics shaped by distinct structural setups. Consequently, only few immature efforts have been made in this direction, typically Pan \textit{et al.}\cite{pan2008transfer} attempting to transfer local knowledge within the same areas using a common model pre-trained over different periods. By extension, Yong \textit{et al.}\cite{yong2020indoor} explored pre-training models on similar areas with upsampled and downsampled input data. More recently, Klus \textit{et al.}\cite{klus2021transfer} adopted separate encoders to create input features of the same size across different RSS datasets for transfer learning. However, because of the holistic pre-training process on heterogeneous source datasets, the improvements on the target datasets remain somewhat limited. 

Notably, our framework distinguishes itself totally from both regular knowledge distillation and transfer-learning paradigms for offering a comprehensive representation alignment. This uniqueness cleverly sidesteps issues associated with forgetting in transfer-learning and the single-sided effects of knowledge distillation.
In contrast to distillation methods that mandate teachers' representations to be derived only from the target dataset for imitation, our framework furnishes reference representations modeled from various source datasets. This provides specialized networks with more degrees of freedom in alignment, contributing to a richer learning process.
Furthermore, unlike typical transfer-learning paradigms that adopt pre-trained networks on individual sources as initial points for the fine-tuning process on the target dataset, our framework, particularly in the Expert Distilling phase, aligns learned representations of untrained networks on the target collectively with essential knowledge from multiple sources, thereby enhancing the interaction in the learning process and utilizability of pre-existing knowledge more radically.
\vspace{-8pt}

\section{Proposed Method}
\vspace{-5pt}
\begin{figure*}[t!]
	\centering
	\includegraphics[width=0.75\linewidth]{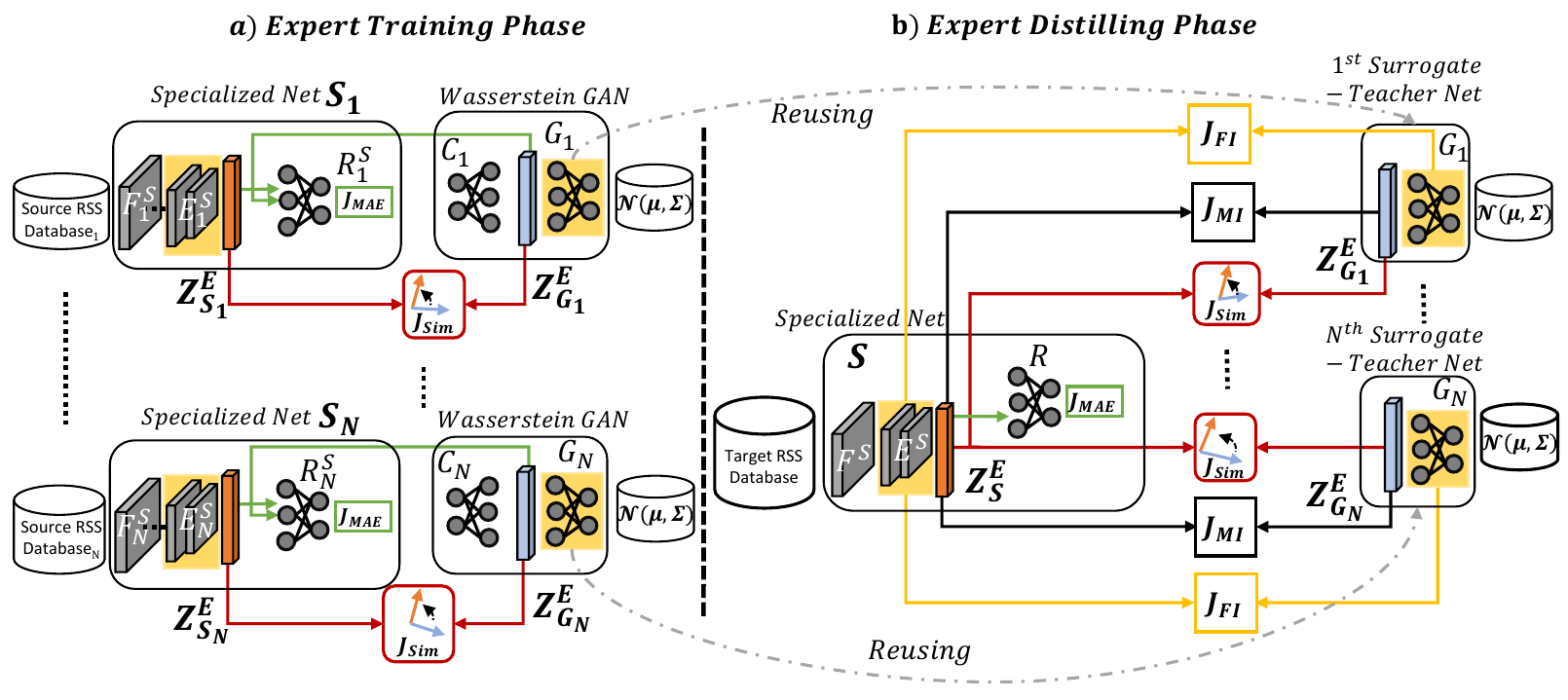}%
	\vspace{-10pt}
	\caption{\textbf{The Proposed Framework}. 
		a) In the \textit{Expert Training} phase, generative models dubbed $G_i$ function as surrogate teachers, alongside associated critics represented by $C_i$, modeling representation spaces established by the specialized network on their respective source datasets. The quality of these modeled features is rigorously assessed by \textbf{Regressors} $R_i^{S}$, a last part of the specialized network $S_i$, and Angular Similarity $J_{Sim}$ (represented by the green and red flows respectively). 
		b) The \textit{Expert Distilling} phase aligns specialized representations learned on the target dataset with those modeled by surrogate teachers. 
	}
	
	\label{fig:overall}
	\vspace{-15pt}
\end{figure*}
Outlined in Fig.\ref{fig:overall}, our framework performs the alignment in two main phases. The \textit{Expert Training} phase equips surrogate-teacher networks to faithfully model the representation spaces formerly founded by specialized networks on their respective source datasets. Following this, the \textit{Expert Distilling} phase focuses on distilling these representations into reference knowledge for alignment.
\vspace{-5pt}
\subsection{Expert Training phase}
\vspace{-2pt}
Technically, for a specialized model {\small${S_i} = \{ {F^S_i,{E^S_i,{R^S_i} }\}, \forall i \in N}$} associated with one of $N$ source RSS fingerprint datasets, we decompose it into three sequential components: \textbf{Framer} $F^S_i$ responsible for crafting features from a batch of $B$ RSS fingerprints, where each fingerprint ${X = [x_0, \cdots ,x_n] \in \mathbb{R}^{1 \times n}}$ comprises RSS measurements from $n$ anchors; \textbf{Extractor} $E^S_i$ built upon the crafted features to produce specialized representations denoted as ${Z_{S_i}^{E}}$; and \textbf{Regressor} $R^S_i$ exploiting these specialized representations for precise location estimation.

\noindent$\triangleright$ \textbf{Surrogate Teachers.} Just like the configurations of the specialized models, we use simple generative models, each consisting of three fully connected layers, represented by {\small$G_i=\{ F^G_i, E^G_i\}$}. These models are employed in the context of WGANs\cite{gulrajani2017improved} with additional gradient penalty enhancements to maintain stability during the modeling process. To bolster further the quality of modeled representations, denoted as {\small${{Z}_{G_i}^{E}}$}, and alleviate information bottlenecks in the subsequent phase, only two extra overarching constraints are imposed to model specialized representations ${Z_{S_i}^{E}}$ from the source datasets with as much generality as possible. This streamlined phase aims to provide the subsequent phase with a greater degree of flexibility in extracting and aligning latent details with the target dataset. 

\noindent$\triangleright$ \textbf{Angular Similarity Constraint.} As described in Eq.\ref{eq:sim1}, we present a constraint denoted as $J_{Sim}$ to control the correlation between the specialized representations {\small$Z_{S_i}^{E}$} and the modeled representations {\small$Z_{G_i}^{E}$}. This constraint incorporates an additional angular margin $\alpha$ of $0.2$ radians to compel the generative models {\small$G_i$} to harder produce feature representations that should closely align with {\small${Z_{S_i}^{E}}$}.
\vspace{-6pt}
\begin{equation}
	\label{eq:sim1}
	\centering
	\scalebox{0.85}{
		$
		{J_{Sim}} = \max \left( {0,1 - cos\left( {arcos\left( {\overline {\widehat {Z}_{S_i}^E\widehat {Z}_{G_i}^E} } \right) + \alpha } \right)} \right)
		$
	}
\end{equation}
where {\scalebox{0.85}{$\widehat {Z}_{S_i}^E = \frac{{Z_{S_i}^E}}{{{{\left\| {Z_{S_i}^E} \right\|}_2}}},\widehat {Z}_{G_i}^E = \frac{{Z_{G_i}^E}}{{{{\left\| {Z_{G_i}^E} \right\|}_2}}},\overline {\widehat {Z}_{S_i}^E\widehat {Z}_{G_i}^E}  = \frac{1}{B}\sum\limits_{k = 1}^B {\widehat {Z}_{k,S_i}^E\widehat {Z}_{k,G_i}^E}$}}. 

\noindent$\triangleright$ \textbf{Mean Absolute Error.} In addition, it is important to ensure that the modeled representations {\small${{Z}_{G_i}^{E}}$} must also contain valuable information for the localization task. We defer to the corresponding $R^S_i$ on this assessment using another constraint, denoted as  $J_{MAE}$, which computes the absolute difference from true locations {\small$Y \in \mathbb{R}{^{B \times 2}}$}, as detailed in Eq.\ref{eq:mae1}. This dual strategy aims to incentivize the model $G_i$ not only to generate comparable representations but also to ensure their fidelity to the true locations.
\vspace{-6pt}
\begin{equation}
	\label{eq:mae1}
	\centering
	\scalebox{0.8}{
		$
		{J_{MAE}} = \frac{1}{B}\sum\limits_{k = 1}^B {\left\| {{Y_k} - {R^S_i}\left( {Z_{k,S_i}^E} \right)} \right\| + \left\| {{Y_k} - {R^S_i}\left( {Z_{k,G_i}^E} \right)} \right\|}
		$
	}
	\vspace{-4pt}
\end{equation}

\vspace{-6pt}
\subsection{Expert Distilling phase}
\vspace{-3pt}
In this phase, we propose a triplet of underlying constraints for optimization that implicitly establishes a comprehensive representational alignment along the way. Specifically, the essential knowledge distilled from the modeled representations $Z_{G_i}^E$ is consulted to align that of representations learned on the target dataset, denoted as $Z_S^E$. The optimization process repeats concurrently between one single model $S$ and a collection of $N$ surrogate teachers $G_i$, which encourages the model to capture transferable representations insensitive to environmental variations, thereby improving its target localization performance.

\noindent$\triangleright$ \textbf{Cross-Mutual Information Maximization Constraint.} Let {\small$p({Z_S^E},Z_{G_i}^E)$} represent  the joint probability density function of specialized representations {\small$Z_S^E$} on the target dataset and modeled representations {\small$Z^{E}_{G_i}$} from surrogate teachers. Similarly, let {\small$p({{Z_S^{{E}}}})$} and  {\small$p({Z^{E}_{Gi}})$} denote the corresponding marginal probability density functions. The mutual information between {\small${{Z_S^{{E}}}}$ and $Z^{E}_{G_i}$} is defined as follows:
\vspace{-6pt}
\begin{equation}
	\label{eq:mi_begin}
	\vspace{-5pt}
	\scalebox{0.85}{$
		\begin{array}{*{20}{c}}
			{I\left( {Z_{G_i}^E,{Z_S^E}} \right) = \int\limits_{Z_{G_i}^E} {\int\limits_{{Z_S^E}} {p\left( {Z_{G_i}^E,{Z_S^E}} \right)} } } 
			{\log \left( {\frac{{p\left( {Z_{G_i}^E,{Z_S^E}} \right)}}{{p\left( {Z_{G_i}^E} \right)p\left( {{Z_S^E}} \right)}}} \right)dZ_{G_i}^Ed{Z_S^E}}
		\end{array}
		$}
	\vspace{1pt}
\end{equation}
The mutual information {
	\small${I\left( {Z_{G_i}^E,{Z_S^E}} \right)}$} in Eq.\ref{eq:mi_begin} can be presented in the form of the Kullback-Leibler divergence between the joint probability distribution $\mathbb{P}({{Z_S^{{E}}}}, {{Z^{E}_{G_i}}})$  and the product of the marginal distributions {\small${\mathbb{P}\left( {{Z_{G_i}^E}} \right)\mathbb{P}\left( {{Z_S^{{E}}}} \right)}$}, \ie, {\scalebox{0.9}{$I\left( {Z_{G_i}^E,{Z_S^E}} \right) = {D_{KL}}\left( {{\mathbb{P}_{Z_{G_i}^E{Z_S^E}}}\left\| {{\mathbb{P}_{Z_{G_i}^E}}{_{{Z_S^E}}}} \right.} \right)$}}. Given that the Jensen-Shannon (JS) divergence is symmetric between these distributions and we desire to maximize mutual information rather than precisely calculate it, we adopt the mutual information estimator\cite{hjelm2018learning} that proved consistent with JS divergence, \ie, {\scalebox{0.7}{$I\left( {{Z^{E}_{G_i}},{Z_S^{{E}}}} \right) = {D_{JS}}\left( {{\mathbb{P}_{{Z^{E}_{G_i}}{Z_S^{{E}}}}}\left\| {{\mathbb{P}_{{Z^{E}_{G_i}}}}{\mathbb{P}_{{Z_S^E}}}} \right.} \right)$}}.
Derived from JS divergence (as detailed in the supplementary material), the cross-mutual information maximization constraint, denoted as $J_{MI}$ in Eq.\ref{eq:mi_final} is presented to maneuver the neural estimator ${\Psi _\theta }:{Z^{E}_{G_i}} \times {Z_S^{{E}}} \to \mathbb{R} $, parameterized by $\theta$. On the contrary to \cite{hjelm2018learning}, our approach instead estimates and maximizes the cross-mutual information between $Z_S^E$ and $Z_{G_i}^{E}$ over $N$ surrogate teachers $G_i$ just using one global estimator $\Psi_\theta$. By maximizing the cross-mutual information between the target and multiple different sources, the model $S$ is enabled to learn adaptive representations proportional to the relevance of each source dataset.
\vspace{-6pt}
\begin{equation}
	\label{eq:mi_final}
	\scalebox{0.9}{
		$
		{{J_{MI}} = \sum\limits_{i = 1}^N \begin{array}{l}
				{\mathbb{E}_{p\left( {Z_{{G_i}}^E,Z_S^E} \right)}}\left[ { - \log \left( {1 + {e^{ - {\Psi _\theta }\left( {Z_{{G_i}}^E,Z_S^E} \right)}}} \right)} \right]\\
				- {\mathbb{E}_{p\left( {Z_{{G_i}}^E} \right)p\left( {Z_S^E} \right)}}\left[ {\log \left( {1 + {e^{{\Psi _\theta }\left( {Z_{{G_i}}^E,Z_S^E} \right)}}} \right)} \right]
		\end{array} }
		$
	}
	\vspace{-5pt}
\end{equation}
Additionally, this encourages the model $S$ to learn only common information among $N$ surrogate teachers $G_i$ while removing exclusive information from specialized representations $Z_S^E$. In order to activate $J_{MI}$ in Eq.\ref{eq:mi_final}, we draw input samples from $\mathbb{P}(Z_S^E, Z_{G_i}^{E})$ by directly extracting $(Z_S^E, Z^{E}_{G_i})$ pairs from Extractors $E^S$ and $E_i^G$, respectively. Also, we shuffle the representations in batch samples taken from $\mathbb{P}(Z_S^{E}, Z_{G_i}^E)$ to mimic input samples from $\mathbb{P}(Z_S^E)\mathbb{P}(Z^{E}_{G_i})$. 

\noindent$\triangleright$ \textbf{Functional Information Constraint.} Assume a neural network {\small$g\left( X \right):{\mathbb{R}^u} \to {\mathbb{R}^v} = \left( {{L_n} \circ  \cdots  \circ {L_1}} \right)\left( X \right)$} is constituted by $n$ neural layers ${L}$, where each layer outputs a feature map $Z_j = z_j({W_j}Z_{j-1} + {b_j})$. Here, $z_j$, $W_j$, and $b_j$ represent the activation function, weight matrix, and bias vector for layer $L_j$, respectively. For simplicity, the neural network might be reduced to {\small$g\left( X \right) = \left( {{W_n} \circ  \cdots  \circ {W_1}} \right)\left( X \right)$}, where the bias terms $\{{b_j}\}_{j=1}^n$ are neglected. The interpretation can also be approximated to CNNs where the kernel size of $w \times h$ is commensurate with $c$ input and $o$ output channels. This results in a parameter matrix of $o \times chw$, which is similar to that of fully connected layers. Now, the function {\small$g = {\mathbb{R}^u} \to {\mathbb{R}^v}$} is considered Lipschitz continuous if there exists a constant $K$ such that:
\vspace{-5pt}
\begin{equation}
	\label{eq:lip_def}
	\small
	\forall {X_1},{X_2} \in {\mathbb{R}^u},{\left\| {g\left( {{X_1}} \right) - g\left( {{X_2}} \right)} \right\|_2} \le K{\left\| {{X_1} - {X_2}} \right\|_2}
	\vspace{-5pt}
\end{equation}
where {\small${X_1}$, ${X_2}$} represent two random inputs of the function $g$. The smallest $K$ value that satisfies the inequality is the Lipschitz constant of the function $g$, denoted as {\small${\left\| g \right\|_{Lip}}$}. According to Eq.\ref{eq:lip_def}, the Lipschitz constant ${\left\| \cdot\right\|_{Lip}}$ can be defined as an upper bound of the ratio between the variation in the output and the given perturbation in the input. Therefore, it naturally serves as an indicator of functional information related to the robustness (\ie \textit{sensitivity}) of a neural network\cite{von2004distance,virmaux2018lipschitz,bartlett2017spectrally} to arbitrary perturbations. 

Due to inherent discrepancies across RSS datasets, specialized models, even when sharing the same architecture, exhibit varied sensitivity. Therefore, this concept is considered for the sensitivity of specialized models to different RSS datasets and plays a crucial role in enhancing resilience against environmental factors. The Lipschitz constant {\small${\left\| {{L_j}} \right\|_{Lip}}$} can be computed through the spectral norm of the weight matrix {\small${\left\| W_j \right\|_{SN}}$} which is its largest singular value, as specified in Eq.\ref{eq:lip_intro}.
\vspace{-8pt}
\begin{equation}
	\label{eq:lip_intro}
	\scalebox{0.9}{
		$
		\begin{array}{l}
			{\left\| {{L_j}} \right\|_{Lip}} = {\sup _{{Z_{j - 1}}}}{\left\| {\nabla {L_j}\left( {{Z_{j - 1}}} \right)} \right\|_{SN}} = {\sup _{{Z_{j - 1}}}}{\left\| {\nabla {Z_j}} \right\|_{SN}}\\
			\hspace{90pt}{\rm{   where}}\\
			{\left\| {\nabla {Z_j}} \right\|_{SN}}  = {\left\| W_j \right\|_{SN}} \buildrel \Delta \over = \mathop {\max }\limits_{{Z_{j - 1}}:{Z_{j - 1}} \ne 0} \frac{{{{\left\| {W_j{Z_{j - 1}}} \right\|}_2}}}{{{{\left\| {{Z_{j - 1}}} \right\|}_2}}} \\= \mathop {\max }\limits_{{{\left\| {{Z_{j - 1}}} \right\|}_2} \le 1} {\left\| {W_j{Z_{j - 1}}} \right\|_2}
		\end{array}
		$
	}
	\vspace{-4pt}
\end{equation}
Formally, most activation functions, \eg, ReLU, Leaky ReLU, Tanh, Sigmoid, and particularly max-pooling, hold a Lipschitz constant of 1. Other layers, including dropout, batch normalization, and other pooling methods, also maintain explicit Lipschitz constants\cite{goodfellow2016deep}. According to the definition\cite{bartlett2017spectrally}, which shows that {\small${\left\| {{L_n} \circ  \cdots  \circ {L_1}} \right\|_{Lip}} \le {\left\| {{L_n}} \right\|_{Lip}} \cdot  \cdot  \cdot {\left\| {{L_1}} \right\|_{Lip}}$}, the bound of {\small${\left\| g \right\|_{Lip}}$} is further elaborated in Eq.\ref{eq:lip_ext}.
\vspace{-15pt}
\begin{equation}
	\label{eq:lip_ext}
	\scalebox{0.85}{
		$
		{\left\| g \right\|_{Lip}} \le {\left\| {{L_n}} \right\|_{Lip}} \cdot  \cdot  \cdot {\left\| {{L_1}} \right\|_{Lip}} = \prod\limits_{j = 1}^L {{{\left\| {{L_j}} \right\|}_{Lip}}}  = \prod\limits_{j = 1}^L {{{\left\| {{W_j}} \right\|}_{SN}}}
		$
	}
	\vspace{-5pt}
\end{equation}
However, computing the exact Lipschitz constant of neural networks is a computationally challenging problem, even when considering the spectral norm {\small${\left\| W_j \right\|_{SN}}$}.
To circumvent the
complicated calculation of the spectral norm {\small${\left\| W_j \right\|_{SN}}$} and harmonize the constraint with the two preceding representation-level constraints, we
indirectly compute a transmitting matrix {\small${T_j}$}\cite{shang2021lipschitz}, as defined by:
\vspace{-7pt}
\begin{equation}
	\vspace{-5pt}
	\label{eq:lip_indir}
	\scalebox{0.8}{
		$
		\begin{array}{c}
			{T_j} \buildrel \Delta \over = {\left[ {{{\left( {\widehat {Z}_{j - 1}} \right)}^\Gamma }\left( \widehat {Z}_j \right)} \right]^\Gamma }\left[ {{{\left( {\widehat {Z}_{j - 1}} \right)}^\Gamma }\left( \widehat {Z}_j \right)} \right]\\
			\forall \widehat {Z} = \frac{{Z}}{{{{\left\| {Z} \right\|}_2}}} \in {\mathbb{R}^{d \times M}}
		\end{array}
		$
	}
	\vspace{-1pt}
\end{equation}
In this equation, {\small${Z}_{j-1}$} and {\small${Z}_{j}$} refer to input and output batches of M $d$-dimensional feature maps before and after the layer $L_j$, respectively. It is observed that a batch of feature maps at the same layer expectantly has strong mutual linear independence for well-trained networks\cite{chen2017exemplar,tung2019similarity}. This implies that {{\scalebox{0.83}{${\left( \widehat {Z}_j \right)^\Gamma }\left( \widehat {Z}_j \right) \approx {\rm I}$}}, where {\small$\rm I$} is a unit matrix. Based on this, {\small${T_j}$} is further derived as follows:
	\vspace{-7pt}
	\begin{equation}
		\vspace{-5pt}
		\label{eq:lip_ext2}
		\scalebox{0.8}{
			$
			\begin{array}{c}
				{T_j} \buildrel \Delta \over = {\left[ {{{\left( {\widehat {Z}_{j - 1}} \right)}^\Gamma }\left( \widehat {Z}_j \right)} \right]^\Gamma }\left[ {{{\left( {\widehat {Z}_{j - 1}} \right)}^\Gamma }\left( \widehat {Z}_j \right)} \right]\\
				= {\widehat {Z}_{j - 1}^\Gamma }\left( {{W_j}^\Gamma {W_j}} \right)\widehat {Z}_{j - 1}
			\end{array}
			$
		}
		\vspace{-5pt}
	\end{equation}
	Since matrix {\small${\widehat {Z}_{j - 1}}$} is an orthogonal matrix, which was expressed by {\scalebox{0.8}{${\left( {\widehat {Z}_{j - 1}} \right)^\Gamma }\left( {\widehat {Z}_{j - 1}} \right) \approx {\rm I}$}}, the largest eigenvalues ${\sigma _1}(\cdot)$
	of {\scalebox{0.83}{
			${\widehat {Z}_{j - 1}^\Gamma }{\left( {{W_i}^\Gamma {W_j}} \right)}\widehat {Z}_{j - 1}$} and {\scalebox{0.83}{${{W_j}^\Gamma {W_j}}$}} should be equivalent, as shown in Eq.\ref{eq:lip_final}.
		\vspace{-8pt}
		\begin{equation}
			\vspace{-5pt}
			\label{eq:lip_final}
			\scalebox{0.85}{
				$
				\begin{array}{c}
					{\sigma _1}\left( {{{\widehat {Z}_{j - 1}}^\Gamma }\left( {{W_j}^\Gamma {W_j}} \right)\widehat {Z}_{j - 1}} \right) = {\sigma _1}\left( {{W_j}^\Gamma {W_j}} \right)\\
					\Leftrightarrow {\sigma _1}\left( {{T_j}} \right) = {\sigma _1}\left( {{W_j}^\Gamma {W_j}} \right) = {\left\| {{W_j}} \right\|_{SN}}
				\end{array}
				$
			}
			\vspace{-1pt}
		\end{equation}
		Thus, the spectral norm {\small${\left\| W_j \right\|_{SN}}$} can be computed by finding the largest eigenvalue of {\small${T_j}$}. This computation of the Lipschitz constant can also be extended to entire blocks of neural networks by simply considering their input and output feature maps, for instance {\small$Z^{F}_{G_i}$} and {\small$Z^{E}_{G_i}$}. In this way, we can transfer the Lipschitz constants of Extractors $E^{G}_i$ in surrogate teachers as functional information to the specialized model $S$ via $L_{FI}$ as depicted in Eq.\ref{eq:J_FI}. This is achieved by utilizing the Power Iteration Method\cite{yoshida2017spectral,shang2021lipschitz}.
		\vspace{-8pt}
		\begin{equation}
			\vspace{-5pt}
			\label{eq:J_FI}
			\scalebox{0.8}{
				$
				\begin{array}{*{20}{c}}
					{{J_{FI}} = \sum\limits_{i = 1}^N {{{\left\| {{{\left\| {E_i^G} \right\|}_{SN}} - {{\left\| {{E^S}} \right\|}_{SN}}} \right\|}_2}} }\\
					{ = \sum\limits_{i = 1}^N {{{\left\| {{\sigma _1}\left( {T\left( {Z_{G_i}^F,Z_{G_i}^E} \right)} \right) - {\sigma _1}\left( {T\left( {{Z_S^F},{Z_S^E}} \right)} \right)} \right\|}_2}} }
				\end{array}
				$
			}
			\vspace{-2pt}
		\end{equation}
		\noindent$\triangleright$ \textbf{Overall Loss.} 
		The overall loss, as described in Eq.\ref{eq:J_overall}, combines all the proposed constraints from this phase. Furthermore, it incorporates weighting factors {\small${\overline\lambda_{t=1}^4}$}, where {\small$\overline {{\lambda _t}}  = \frac{{{\lambda _t}}}{{\sum\limits_{k = 1}^4 {{\lambda _k}} }}$}, to balance the influence of different types of knowledge in the loss function.
		\vspace{-5pt}
		\begin{equation}
			\label{eq:J_overall}
			\small
			{J_{overall}} = {\overline\lambda _1}{J_{MAE}} + {\overline\lambda _2}{J_{Sim}} + {\overline\lambda _3}{J_{MI}} + {\overline\lambda _4}{J_{FI}}
			\vspace{-5pt}   
		\end{equation}
		\section{Experiments}
		\vspace{-3pt}
		\subsection{Implementation Details}
		\vspace{-4pt}
		\textbf{Databases.} 
		Evaluation is conducted on three distinct benchmark RSS fingerprint databases: UJIIndoorLoc\cite{torres2014ujiindoorloc}, UTS\cite{song2019cnnloc}, and Tampere\cite{lohan2017wi}. These large-scale databases exhibit distinct characteristics, including differences in the input number and arrangement of WiFi anchors, types of capturing devices, buildings, inner infrastructures, and layouts of surrounding subjects. Notably, there is a significant level of sparsity in these databases, with only a fraction of the available WiFi anchors being operational for each fingerprint.
		For example, in UJIIndoorLoc, approximately $232$ out of $520$ anchors on each floor were active.  In UTS and Tampere, $557$ out of $589$, and $779$ out of $992$  anchors were actually in operation, respectively.  The missing anchor values were filled with default values of  $100$ $dB$. This sparsity poses a challenge, particularly in fine-grained localization scenarios that demand detailed information from a wide range of anchors to achieve accurate results.
		
		\noindent\textbf{Evaluation Metrics.} 
		For a rigorous comparison, we employ a fine-grained evaluation approach using the Mean Absolute Error (MAE) metric to assess the overall localization performance. The MAE is calculated by measuring the Euclidean distance between estimated locations and their corresponding ground truth. Additionally, we consider the $75^{th}$ Percentile and $95^{th}$ Percentile for stability examination.
		
		\noindent\textbf{Parameter Setting.} 
		In our experiments, we set the weighting factors $\lambda_{i=1}^4$ to $3.0$, $0.5$, $0.5$, and $0.5$ in sequence, determined through a grid search on UJIIndoorLoc. All phases of our experiments are implemented in TensorFlow 2.11, using a fixed batch size of $B=128$. The models are trained for 2000, 500, and 400 epochs at learning rates of $1e-3$, $1e-3$, and $1e-4$ for Tampere, UJIIndoorLoc, and UTS datasets, respectively. These experiments utilized Adam optimization and were conducted on a single NVIDIA A100 GPU with 40GB of memory.
		\vspace{-5pt}
		\subsection{Comparisons to the State-of-the-art Methods}
		\label{Subsec:comparison}
		\vspace{-5pt}
	\begin{table*}[t!]
		\caption{Comparison of SoTA specialized networks  in indoor localization}
		\label{tab:comp}
		\vspace{-9pt}
		\centering
		\resizebox{1.6\columnwidth}{!}{%
			\begin{tabular}{lccccccccccc}
				\hline
				\multicolumn{1}{c}{\multirow{3}{*}{\textbf{Method}}} &
				\multicolumn{3}{c}{\textbf{UJIIndoorLoc}} &
				\multicolumn{1}{l}{} &
				\multicolumn{3}{c}{\textbf{UTS}} &
				\multicolumn{1}{l}{} &
				\multicolumn{1}{l}{} &
				\textbf{Tampere} &
				\textbf{} \\ \cline{2-4} \cline{6-8} \cline{10-12} 
				\multicolumn{1}{c}{} &
				\multirow{2}{*}{MAE (m)$\downarrow$} &
				\multicolumn{2}{c}{Percentile (m) $\downarrow$} &
				\multicolumn{1}{l}{} &
				\multirow{2}{*}{MAE (m)$\downarrow$} &
				\multicolumn{2}{c}{Percentile (m) $\downarrow$} &
				\multicolumn{1}{l}{} &
				\multirow{2}{*}{MAE (m)$\downarrow$} &
				\multicolumn{2}{c}{Percentile (m) $\downarrow$} \\ \cline{3-4} \cline{7-8} \cline{11-12} 
				\multicolumn{1}{c}{} &
				&
				$75^{th}$ &
				$95^{th}$ &
				\multicolumn{1}{l}{} &
				&
				$75^{th}$ &
				$95^{th}$ &
				\multicolumn{1}{l}{} &
				&
				$75^{th}$ &
				$95^{th}$ \\ \hline
				DNN\cite{felix2016fingerprinting} &
				14.04 &
				18.37 &
				35.99 &
				&
				7.69 &
				11.20 &
				18.86 &
				&
				11.03 &
				13.59 &
				24.31 \\
				DNN* &
				16.05 (14.32$\%\downarrow$)&
				21.20 &
				39.10 &
				&
				61.29 (697.01$\%\downarrow$)&
				76.78 &
				89.53 &
				&
				12.36 (12.06$\%\downarrow$)&
				14.75 &
				34.31 \\
				DNN**\cite{klus2021transfer} &
				21.62 (53.99$\%\downarrow$)&
				29.12&
				48.15 &
				&
				9.58 (24.58$\%\downarrow$)&
				12.98 &
				19.66 &
				&
				11.84 (7.34$\%\downarrow$)&
				15.54 &
				26.53 \\
				\textbf{DNN+++} &
				\textbf{11.98 (14.67$\%\uparrow$)} &
				\textbf{14.92} &
				\textbf{34.08} &
				&
				\textbf{6.68 (13.13$\%\uparrow$)} &
				\textbf{10.79} &
				\textbf{16.88} &
				&
				\textbf{8.86 (19.67$\%\uparrow$)} &
				\textbf{11.30} &
				\textbf{21.57} \\ \hline
				CNNLoc\cite{song2019novel} &
				13.12 &
				17.30 &
				34.80 &
				&
				7.19 &
				11.54 &
				18.70 &
				&
				10.06 &
				12.87 &
				23.70 \\
				CNNLoc* &
				15.32 (16.77$\%\downarrow$)&
				19.53 &
				38.59 &
				&
				7.79 (8.35$\%\downarrow$)&
				10.14 &
				17.25 &
				&
				12.66 (25.85$\%\downarrow$)&
				15.46 &
				33.76 \\
				CNNLoc**\cite{klus2021transfer} &
				27.71 (111.20$\%\downarrow$)&
				37.09 &
				62.94 &
				&
				9.59 (33.38$\%\downarrow$)&
				12.46 &
				20.51 &
				&
				11.36 (12.92$\%\downarrow$)&
				14.47 &
				25.73 \\
				\textbf{CNNLoc+++} &
				\textbf{11.67 (11.05$\%\uparrow$)} &
				\textbf{15.14} &
				\textbf{32.69} &
				&
				\textbf{6.80 (5.42$\%\uparrow$)} &
				\textbf{10.74} &
				\textbf{16.79} &
				&
				\textbf{8.25 (17.99$\%\uparrow$)} &
				\textbf{10.31} &
				\textbf{20.32} \\ \hline
				BayesCNN\cite{sinha2021completely} &
				12.63 &
				16.62 &
				33.40 &
				&
				8.19 &
				11.89 &
				18.31 &
				&
				10.31 &
				13.10 &
				23.71 \\
				BayesCNN* &
				15.07 (19.32$\%\downarrow$)&
				19.91 &
				35.57 &
				&
				8.10 (1.10$\%\uparrow$)&
				\textbf{10.57} &
				17.21 &
				&
				11.25 (9.12$\%\downarrow$)&
				13.76 &
				29.51 \\
				BayesCNN**\cite{klus2021transfer} &
				129.95 (929.00$\%\downarrow$)&
				196.66 &
				225.14 &
				&
				9.48 (15.75$\%\downarrow$)&
				12.85 &
				20.50 &
				&
				12.06 (16.97$\%\downarrow$)&
				15.67 &
				28.23 \\
				\textbf{BayesCNN+++} &
				\textbf{10.79 (14.57$\%\uparrow$)} &
				\textbf{14.03} &
				\textbf{29.45} &
				&
				\textbf{6.62 (19.17$\%\uparrow$)} &
				10.67 &
				\textbf{17.71} &
				&
				\textbf{9.01 (12.67$\%\uparrow$)} &
				\textbf{11.50} &
				\textbf{22.74} \\ \hline
				bAaT\cite{nguyen2023learning} &
				11.51 &
				15.18 &
				\textbf{31.24} &
				&
				6.38 &
				9.87 &
				\textbf{15.32} &
				&
				7.99 &
				10.08 &
				19.89 \\
				bAaT* &
				13.46 (16.95$\%\downarrow$)&
				17.73 &
				36.44 &
				&
				7.91 (23.98$\%\downarrow$)&
				10.07 &
				16.47 &
				&
				10.77 (34.79$\%\downarrow$)&
				13.15 &
				26.91 \\
				bAaT**\cite{klus2021transfer} &
				98.86 (758.91$\%\downarrow$)&
				146.3 &
				179.25 &
				&
				8.70 (36.36$\%\downarrow$)&
				11.30 &
				18.85 &
				&
				10.63 (33.04$\%\downarrow$)&
				13.18 &
				25.35 \\
				\textbf{bAaT++}(wo $J_{FI}$) &
				\textbf{11.15 (3.13$\%\uparrow$)} &
				\textbf{14.51} &
				32.56 &
				&
				\textbf{6.30 (1.25$\%\uparrow$)} &
				\textbf{9.78} &
				15.57 &
				&
				\textbf{7.73 (3.25$\%\uparrow$)} &
				\textbf{10.00} &
				\textbf{19.54} \\ \hline
			\end{tabular}%
		}
		\vspace{-5pt}
		\begin{flushleft}
			\footnotesize{Notably, the combinations of source and target datasets are denoted as follows: \textit{UTS}$\&$\textit{Tampere}$\to$\textit{UJI}, \textit{UJI}$\&$\textit{Tampere}$\to$\textit{UTS}, and \textit{UJI}$\&$\textit{UTS}$\to$\textit{Tampere}.}
			
			\footnotesize{(*): Sequential Transfer-Learning using Encoder networks.}
			
			\footnotesize{(**): Collective Transfer-Learning using Encoder networks\cite{klus2021transfer}.}
			
			\footnotesize{
				(\textbf{+++}): Fully enhanced versions of the specialized models that have received assistance from the proposed framework for knowledge transfer.}
			
			\footnotesize{\textbf{(++}): Because of Lipschitz discontinuity in Transformer-based architectures\cite{kim2021lipschitz}, bAaT++ is merely enhanced with the first two constraints, but $J_{FI}$.} 
		\end{flushleft}
		\vspace{-18pt}
	\end{table*}
	The proposed framework delivers a significant overall performance enhancement for stand-alone models, as measured by metrics including MAE, the $75^{th}$ Percentile, and the $95^{th}$ Percentile. As shown in Tab.\ref{tab:comp}, most specialized models benefit considerably from the framework across all target datasets. Interestingly, the extent of performance improvement appears to be correlated with the architectural complexity of the models. For example, in UJIIndoor, the simple DNN+++ model in UJIIndoorLoc experiences a notable reduction of approximately 2.06 meters, translating to a 14.67$\%$ improvement in MAE, while more complex models like CNNLoc+++ and BayesCNN+++ show smaller improvements, with maximum MAE reductions of 1.45 meters (11.05$\%$), and 1.84 meters (\ie 14.57$\%$) respectively. 
	The differences might also be influenced by the scale and characteristics of the target datasets, as fully enhanced versions achieve marginal improvements in UTS, where RSS fingerprints were collected in smaller areas.
	
	Furthermore, the framework is benchmarked against other transfer-learning methods designed for fingerprint-based indoor localization. Some of these methods yield substantial estimation errors, exemplified by 61.29 meters and 129.95 meters MAEs for DNN* and BayesCNN** in UTS and UJIIndoorLoc, respectively. These failures can be attributed to the use of basic transfer-learning schemes (\eg, sequential and holistic transfer), which employ irrelevant representations learned by their encoder networks, thus constraining the learning capability of SoTA methods. In contrast, the proposed framework, dynamically empowering specialized models with ample references for alignment, exhibits overwhelming superiority with a considerable MAE reduction ranging from approximately 11$\%$ to 20$\%$, depending on architectures.
	
	For stability evaluation, it is observed that $75\%$ of the estimated locations generated by all fully enhanced versions show significantly lower distance errors compared to both the baseline and other transfer-learning methods across all target datasets. Moreover, the accumulative error plots in Fig.\ref{fig:cdf} also provide a comprehensive overview of the overall performance improvements across SoTA specialized models. In particular, simple models, namely DNN+++, and CNNLoc+++ can easily reach, and sometimes even surpass, the performance of complex architectures, particularly bAaT, over the target datasets. This is indicated by the solid lines located above the orange dashed line.
	
	Looking more closely at Fig.\ref{fig:cdf}, these fully enhanced versions consistently outperform their respective stand-alone counterparts, with the most noticeable improvements occurring in the range from the $15^{th}$ Percentile to the $75^{th}$ Percentile, described by the upper solid curves in UJIIndoorLoc and Tampere. In addition, the long-term stability improvements of the proposed framework are also emphasized by a substantial increase in the accumulative distribution function of SoTA specialized models. For example, the likelihood of the variable $X$ being less than or equal to $10\,m$, represented by the probability $P(X<10\,m)$, for all specialized models experiences a significant rise from an average of around $50\%$, $\sim60\%$, and $\sim60\%$ to approximately  $62\%$, $\sim72\%$, and $\sim75\%$ for UJIIndoor, UTS, and Tampere, respectively. This reliability boost positions these networks for more practical and dependable applications.

	{\setlength{\tabcolsep}{0.5pt}
		\begin{figure*}[t!]
			\vspace{-2.5pt}
			\hspace{-5pt}{
				\begin{tabular}{>{\raggedleft}p{0.34\textwidth}p{0.34\textwidth}>{\raggedleft\arraybackslash}p{0.34\textwidth}}
					\includegraphics[width=.34\textwidth] {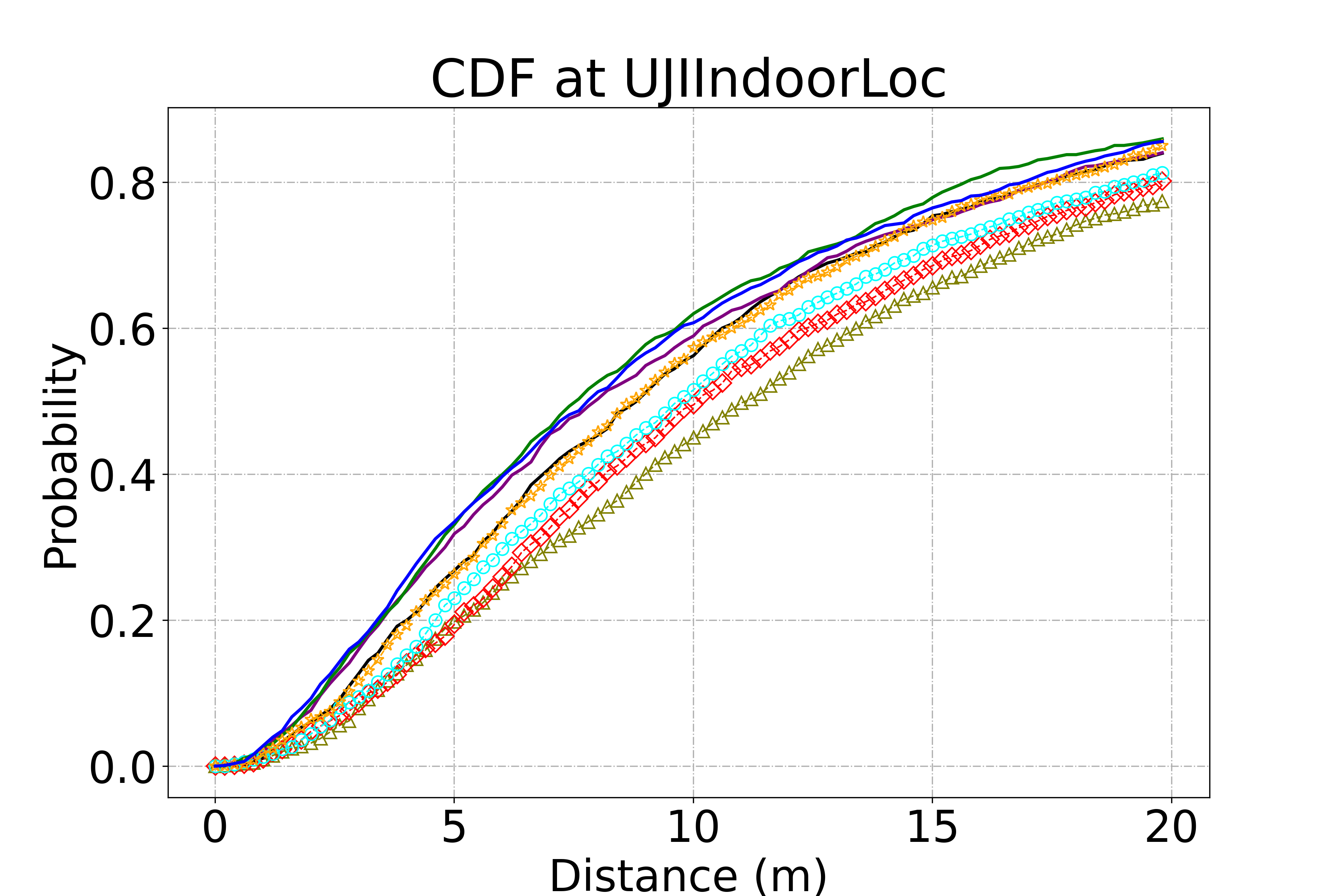} &
					\includegraphics[width=.34\textwidth]{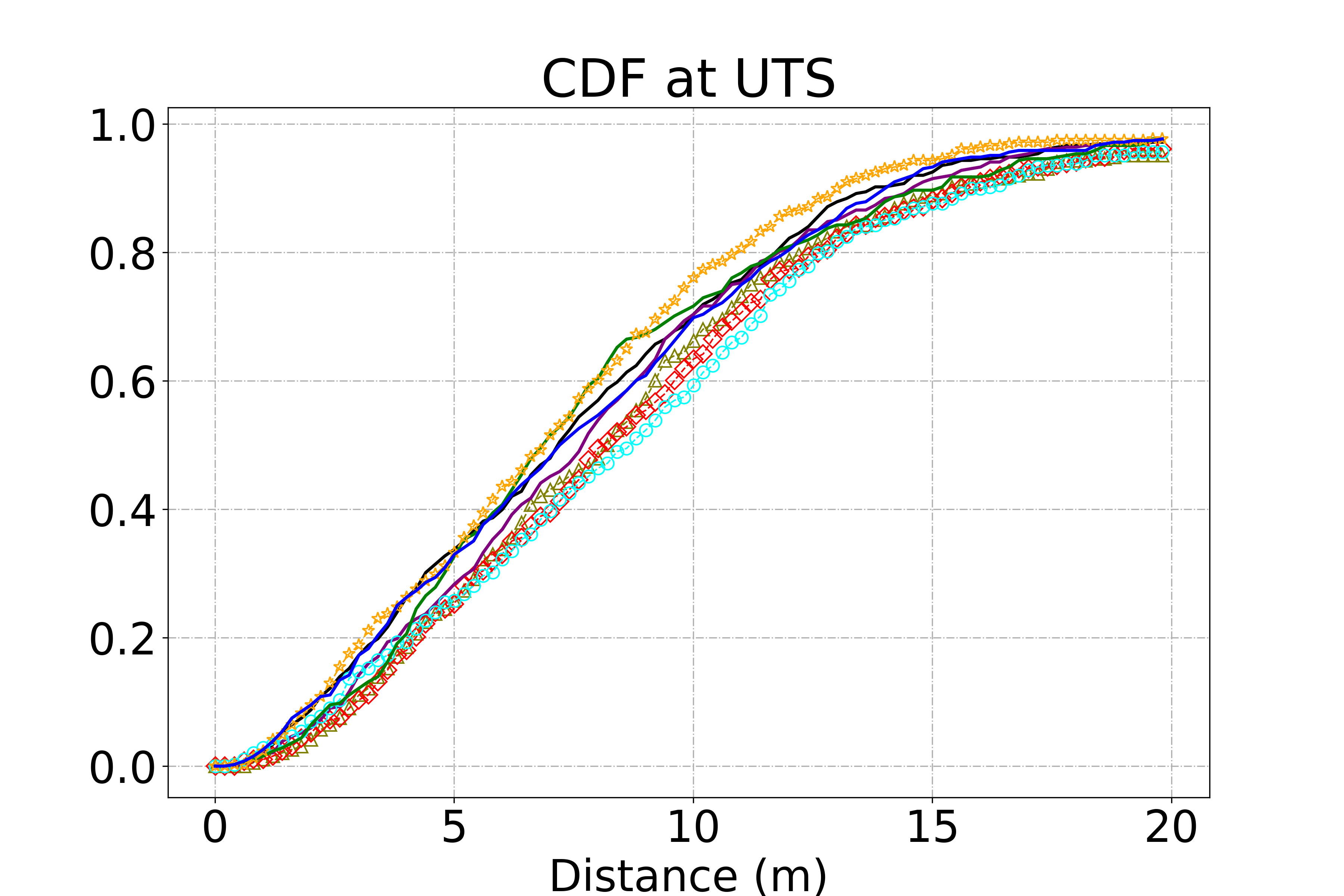} &
					\includegraphics[width=.34\textwidth]{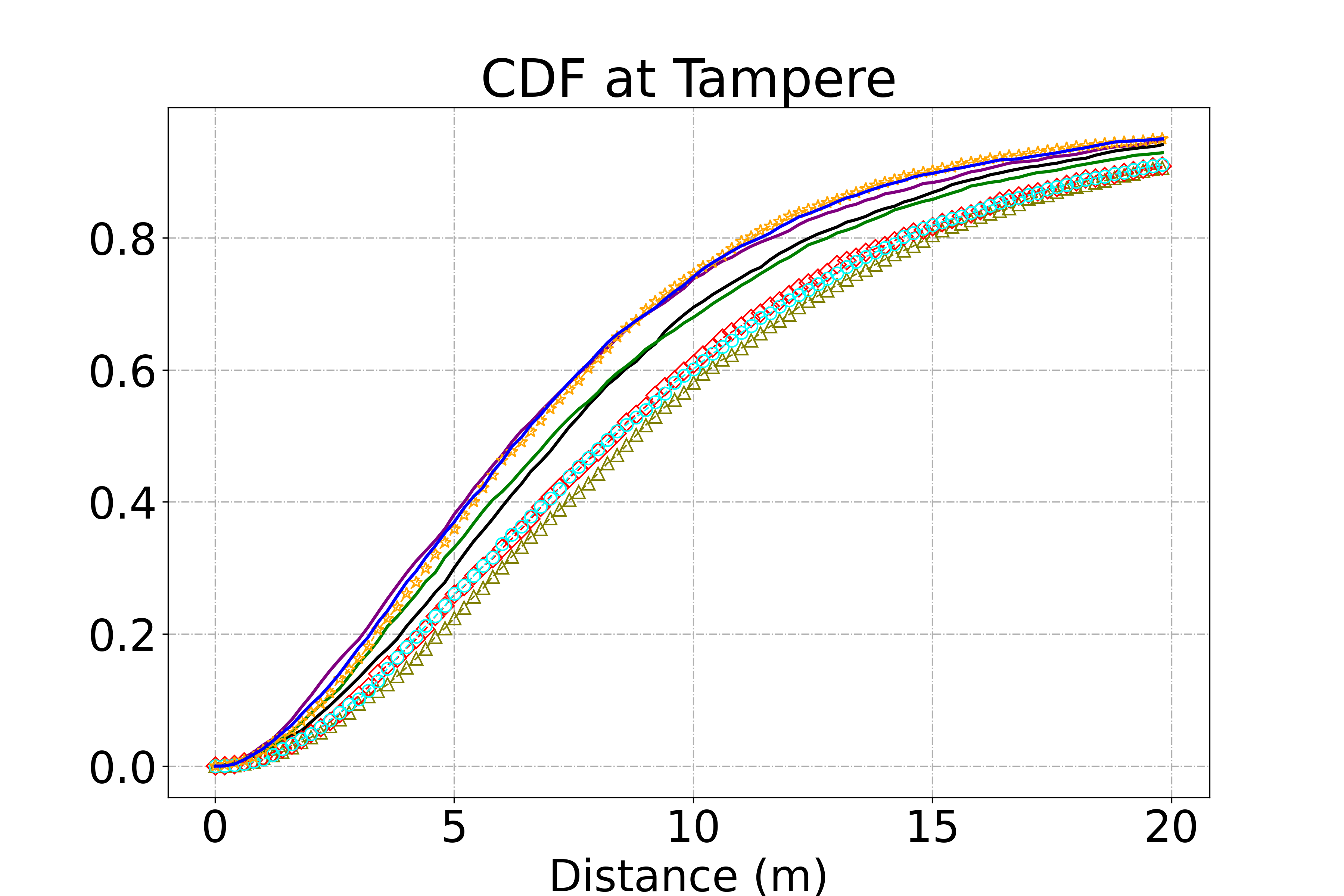}\\
				\end{tabular}
			}
			\vspace{-10pt}
			\centering
			\includegraphics[width=0.9\textwidth]{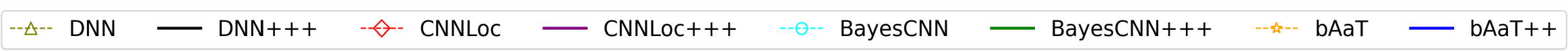}
			\caption{Empirical Cumulative Distribution Function of the framework applied to state-of-the-art models on three indoor localization datasets. The solid curves represent the cumulative errors made by the fully enhanced versions, while the dashed curves represent the original models. Best viewed in color and full-screen mode.}
			\label{fig:cdf}
			\vspace{-10pt}
		\end{figure*}
	}
	
	\vspace{-4pt}
	\subsection{Ablation Study}
	\label{subsec:abla}
	\vspace{-4pt}
	
	\begin{table*}[t!]
		\caption{The impact of incorporating the proposed unsupervised constraints on different networks in the Expert Distilling phase}
		\label{tab:ablation}
		\vspace{-9pt}
		\centering
		\resizebox{2.1\columnwidth}{!}{%
			\begin{tabular}{lllllllllllllllllll}
				\cline{1-9} \cline{11-19}
				\multicolumn{1}{c}{\multirow{2}{*}{\textbf{Method}}} &
				\multicolumn{1}{c}{\multirow{2}{*}{$J_{Sim}$}} &
				\multicolumn{1}{c}{\multirow{2}{*}{$J_{MI}$}} &
				\multicolumn{1}{c}{\multirow{2}{*}{$J_{FI}$}} &
				\multicolumn{1}{c}{\textbf{UJIIndoorLoc}} &
				&
				\multicolumn{1}{c}{\textbf{UTS}} &
				&
				\multicolumn{1}{c}{\textbf{Tampere}} &
				&
				\multicolumn{1}{c}{\multirow{2}{*}{\textbf{Method}}} &
				\multicolumn{1}{c}{\multirow{2}{*}{$J_{Sim}$}} &
				\multicolumn{1}{c}{\multirow{2}{*}{$J_{MI}$}} &
				\multicolumn{1}{c}{\multirow{2}{*}{$J_{FI}$}} &
				\multicolumn{1}{c}{\textbf{UJIIndoorLoc}} &
				&
				\multicolumn{1}{c}{\textbf{UTS}} &
				&
				\multicolumn{1}{c}{\textbf{Tampere}} \\ \cline{5-5} \cline{7-7} \cline{9-9} \cline{15-15} \cline{17-17} \cline{19-19} 
				\multicolumn{1}{c}{} &
				\multicolumn{1}{c}{} &
				\multicolumn{1}{c}{} &
				\multicolumn{1}{c}{} &
				\multicolumn{1}{c}{MAE(m)$\downarrow$} &
				&
				\multicolumn{1}{c}{MAE(m)$\downarrow$} &
				&
				\multicolumn{1}{c}{MAE(m)$\downarrow$} &
				&
				\multicolumn{1}{c}{} &
				\multicolumn{1}{c}{} &
				\multicolumn{1}{c}{} &
				\multicolumn{1}{c}{} &
				\multicolumn{1}{c}{MAE(m)$\downarrow$} &
				&
				\multicolumn{1}{c}{MAE(m)$\downarrow$} &
				&
				\multicolumn{1}{c}{MAE(m)$\downarrow$} \\ \cline{1-9} \cline{11-19} 
				DNN\cite{felix2016fingerprinting} &
				&
				&
				&
				14.04 &
				&
				7.69 &
				&
				11.03 &
				&
				BayesCNN\cite{sinha2021completely} &
				&
				&
				&
				12.63 &
				&
				8.19 &
				&
				10.31 \\
				DNN+ &
				$\checkmark$ &
				&
				&
				13.98 (0.43$\%\uparrow$)&
				&
				7.70 (0.13$\%\downarrow$)&
				&
				10.14 (8.07$\%\uparrow$)&
				&
				BayesCNN+ &
				$\checkmark$ &
				&
				&
				\cellcolor{red!15}10.86 (14.01$\%\uparrow$)&
				&
				\cellcolor{red!15}7.46 (8.91$\%\uparrow$)&
				&
				9.97 (3.30$\%\uparrow$)\\
				DNN$^\star$ &
				&
				$\checkmark$ &
				&
				\cellcolor{red!15}11.82 (15.81$\%\uparrow$)&
				&
				7.53 (2.08$\%\uparrow$)&
				&
				\cellcolor{red!15}8.71 (21.03$\%\uparrow$)&
				&
				BayesCNN$^\star$ &
				&
				$\checkmark$ &
				&
				10.92 (13.54$\%\uparrow$)&
				&
				7.81 (4.64$\%\uparrow$)&
				&
				9.19 (10.86$\%\uparrow$)\\
				DNN++ &
				$\checkmark$ &
				$\checkmark$ &
				&
				\textbf{\cellcolor{red!25}11.70} (16.67$\%\uparrow$)&
				&
				\cellcolor{red!15}7.09 (7.80$\%\uparrow$)&
				&
				\textbf{\cellcolor{red!25}8.39 (23.94$\%\uparrow$)} &
				&
				BayesCNN++ &
				$\checkmark$ &
				$\checkmark$ &
				&
				10.95 (13.30$\%\uparrow$)&
				&
				7.76 (5.25$\%\uparrow$)&
				&
				\cellcolor{red!5}9.14 (11.35$\%\uparrow$)\\
				DNN$^\diamond$ &
				&
				&
				$\checkmark$ &
				12.51 (10.90$\%\uparrow$)&
				&
				\cellcolor{red!5}7.35 (4.42$\%\uparrow$)&
				&
				9.81 (11.06$\%\uparrow$)&
				&
				BayesCNN$^\diamond$ &
				&
				&
				$\checkmark$ &
				\cellcolor{red!5}10.87 (13.94$\%\uparrow$)&
				&
				\cellcolor{red!5}7.74 (5.50$\%\uparrow$)&
				&
				\cellcolor{red!15}9.13 (11.45$\%\uparrow$)\\
				DNN+++ &
				$\checkmark$ &
				$\checkmark$ &
				$\checkmark$ &
				\cellcolor{red!5}11.98 (14.67$\%\uparrow$)&
				&
				\textbf{\cellcolor{red!25}6.68 (13.13$\%\uparrow$)} &
				&
				\cellcolor{red!5}8.86 (19.67$\%\uparrow$)&
				&
				BayesCNN+++ &
				$\checkmark$ &
				$\checkmark$ &
				$\checkmark$ &
				\textbf{\cellcolor{red!25}10.79 (14.57$\%\uparrow$)} &
				&
				\textbf{\cellcolor{red!25}6.62 (19.17$\%\uparrow$)} &
				&
				\textbf{\cellcolor{red!25}9.01 (12.61$\%\uparrow$)} \\ \cline{1-9} \cline{11-19} 
				CNNLoc\cite{song2019novel} &
				&
				&
				&
				13.12 &
				&
				7.19 &
				&
				10.06 &
				&
				bAaT\cite{nguyen2023learning} &
				&
				&
				&
				11.51 &
				&
				\cellcolor{red!5}6.38 &
				&
				7.99 \\
				CNNLoc+ &
				$\checkmark$ &
				&
				&
				12.96 (1.22$\%\uparrow$)&
				&
				7.10 (1.25$\%\uparrow$)&
				&
				9.83 (2.29$\%\uparrow$)&
				&
				bAaT+ &
				$\checkmark$ &
				&
				&
				\cellcolor{red!5}11.27 (2.09$\%\uparrow$)&
				&
				\cellcolor{red!15}6.36 (0.31$\%\uparrow$)&
				&
				8.05 (0.75$\%\downarrow$)\\
				CNNLoc$^\star$ &
				&
				$\checkmark$ &
				&
				\cellcolor{red!5}12.72 (3.05$\%\uparrow$)&
				&
				\cellcolor{red!15}6.92 (3.76$\%\uparrow$)&
				&
				\cellcolor{red!15}7.74 (23.06$\%\uparrow$)&
				&
				bAaT$^\star$ &
				&
				$\checkmark$ &
				&
				\cellcolor{red!15}11.24 (2.35$\%\uparrow$)&
				&
				\textbf{\cellcolor{red!25}6.30 (1.25$\%\uparrow$)} &
				&
				\cellcolor{red!15}7.83 (2.00$\%\uparrow$)\\
				CNNLoc++ &
				$\checkmark$ &
				$\checkmark$ &
				&
				12.75 (2.82$\%\uparrow$)&
				&
				7.09 (1.39$\%\uparrow$)&
				&
				\textbf{\cellcolor{red!25}7.72 (23.26$\%\uparrow$)} &
				&
				bAaT++ &
				$\checkmark$ &
				$\checkmark$ &
				&
				\textbf{\cellcolor{red!25}11.15 (3.13$\%\uparrow$)} &
				&
				\textbf{\cellcolor{red!25}6.30 (1.25$\%\uparrow$)} &
				&
				\textbf{\cellcolor{red!25}7.73 (3.25$\%\uparrow$)} \\
				CNNLoc$^\diamond$ &
				&
				&
				$\checkmark$ &
				\cellcolor{red!15}12.12 (7.62$\%\uparrow$)&
				&
				\cellcolor{red!5}6.98 (2.92$\%\uparrow$)&
				&
				8.11 (19.38$\%\uparrow$)&
				&
				bAaT$^\diamond$ &
				&
				&
				$\checkmark$ &
				14.07 (22.24$\%\downarrow$)&
				&
				6.88 (7.84$\%\downarrow$)&
				&
				13.78 (72.47$\%\downarrow$)
				\\
				CNNLoc+++ &
				$\checkmark$ &
				$\checkmark$ &
				$\checkmark$ &
				\textbf{\cellcolor{red!25}11.67 (11.05$\%\uparrow$)} &
				&
				\textbf{\cellcolor{red!25}6.80 (5.42$\%\uparrow$)} &
				&
				\cellcolor{red!5}8.04 (20.08$\%\uparrow$)&
				&
				bAaT+++ &
				$\checkmark$ &
				$\checkmark$ &
				$\checkmark$ &
				12.53 (8.86$\%\downarrow$)&
				&
				6.66 (4.39$\%\uparrow$)&
				&
				\cellcolor{red!5}7.94 (0.63$\%\uparrow$)
				\\ 
				\cline{1-9} \cline{11-19} 
			\end{tabular}%
		}
		\vspace{-15pt}
	\end{table*}

	The ablation analysis, as presented in Tab.\ref{tab:ablation}, independently and incrementally adds each of the proposed constraints to the framework to assess their impact on the overall performance of SoTA specialized models.
	
	\noindent$\triangleright$ \textbf{Effect of Angular Similarity.} The impact of $J_{Sim}$ on reducing the Mean Absolute Error (MAE) is noticeable but is not consistently applied to all cases. In some instances, such as with stand-alone DNN and bAaT models in UTS, and Tampere respectively, the partly enhanced versions, namely DNN+ and bAaT+ experience an increase in MAE. This could be due to the possibility that one single constraint alone might not provide enough informative cues for achieving effective representation alignment. It suggests that employing more constraints might be necessary to filter out irrelevant information and improve alignment.
	
	\noindent$\triangleright$ \textbf{Effect of Mutual Information.} 
	\begin{table*}[h!]
		\caption{The impact of the relevance of data sources in Expert Distilling phase}
		\label{tab:util}
		\vspace{-9pt}
		\centering
		\resizebox{1.5\columnwidth}{!}{%
			\begin{tabular}{llccclccclccc}
				\hline
				\multicolumn{1}{c}{\multirow{2}{*}{\textbf{Method}}} &
				&
				\multicolumn{3}{c}{\textbf{UJIIndoorLoc}} &
				&
				\multicolumn{3}{c}{\textbf{UTS}} &
				&
				\multicolumn{3}{c}{\textbf{Tampere}} \\ \cline{3-5} \cline{7-9} \cline{11-13} 
				\multicolumn{1}{c}{} &
				&
				\multicolumn{1}{c}{UTS} &
				\multicolumn{1}{c}{Tampere} &
				MAE (m) $\downarrow$&
				&
				UJIIndoorLoc &
				Tampere &
				MAE (m) $\downarrow$&
				&
				UJIIndoorLoc &
				UTS &
				MAE (m) $\downarrow$\\ \hline
				DNN\cite{felix2016fingerprinting}        &  & \multicolumn{1}{c}{}   & \multicolumn{1}{c}{}   & 14.04          &  &    &    & 7.69          &  &    &    & 11.03         \\
				DNN$^\dagger$      &  & \multicolumn{1}{c}{\checkmark} & \multicolumn{1}{c}{}   & 13.16 (6.27$\%\uparrow$)         &  & \checkmark &    & 6.98 (9.23$\%\uparrow$)            &  & \checkmark &    & 10.75 (2.54$\%\uparrow$)        \\
				DNN$^\ddagger$      &  & \multicolumn{1}{c}{}   & \multicolumn{1}{c}{\checkmark} & \textbf{11.73 (16.45$\%\uparrow$)} &  &    & \checkmark & 6.69 (13.00$\%\uparrow$)            &  &    & \checkmark & 9.65 (12.51$\%\uparrow$)          \\
				DNN+++ &
				&
				\multicolumn{1}{c}{\checkmark} &
				\multicolumn{1}{c}{\checkmark} &
				11.98 (14.67$\%\uparrow$)&
				&
				\checkmark &
				\checkmark &
				\textbf{6.68 (13.13$\%\uparrow$)} &
				&
				\checkmark &
				\checkmark &
				\textbf{8.86 (19.67$\%\uparrow$)} \\ \hline
				CNNLoc\cite{song2019novel}     &  & \multicolumn{1}{c}{}   & \multicolumn{1}{c}{}   & 13.12          &  &    &    & 7.19          &  &    &    & 10.06         \\
				CNNLoc$^\dagger$   &  & \multicolumn{1}{c}{\checkmark} & \multicolumn{1}{c}{}   & 12.00 (8.54$\%\uparrow$)         &  & \checkmark &    & 7.26 (0.97$\%\downarrow$)         &  & \checkmark &    & 8.24 (18.09$\%\uparrow$)         \\
				CNNLoc$^\ddagger$   &  &                        & \checkmark                     & 12.05 (8.16$\%\uparrow$)         &  &    & \checkmark & 7.70 (7.09$\%\downarrow$)         &  &    & \checkmark & \textbf{7.97 (20.78$\%\uparrow$)} \\
				CNNLoc+++   &  & \checkmark                     & \checkmark                     & \textbf{11.67 (11.05$\%\uparrow$)} &  & \checkmark & \checkmark & \textbf{6.80 (5.42$\%\uparrow$)} &  & \checkmark & \checkmark & 8.04 (20.08$\%\uparrow$)          \\ \hline
				BayesCNN\cite{sinha2021completely}   &  &                        &                        & 12.63          &  &    &    & 8.19          &  &    &    & 10.31         \\
				BayesCNN$^\dagger$ &  & \checkmark                     &                        & 13.17 (4.28$\%\downarrow$)         &  & \checkmark &    & 7.90 (3.54$\%\uparrow$)         &  & \checkmark &    & 10.01 (2.91$\%\uparrow$)          \\
				BayesCNN$^\ddagger$ &  &                        & \checkmark                     & 11.20 (11.32$\%\uparrow$)         &  &    & \checkmark & 8.06 (1.59$\%\uparrow$)         &  &    & \checkmark & 9.59 (6.98$\%\uparrow$)         \\
				BayesCNN+++ &  & \checkmark                     & \checkmark                     & \textbf{10.79 (14.57$\%\uparrow$)} &  & \checkmark & \checkmark & \textbf{6,62 (19.17$\%\uparrow$)} &  & \checkmark & \checkmark & \textbf{8.95 (13.19$\%\uparrow$)} \\ \hline
				bAaT\cite{nguyen2023learning}       &  &                        &                        & 11.51 &  &    &    & 6.38 &  &    &    & 7.99          \\
				bAaT$^\dagger$     &  & \checkmark                     &                        & 20.39 (77.15$\%\downarrow$)             &  & \checkmark &    & 6.82 (6.90$\%\downarrow$)         &  & \checkmark &    & 7.92 (0.88$\%\uparrow$)         \\
				bAaT$^\ddagger$     &  &                        & \checkmark                     & 11.99 (4.17$\%\downarrow$)&  &    & \checkmark & 6.78 (6.27$\%\downarrow$)         &  &    & \checkmark & 9.17 (14.77$\%\downarrow$)         \\
				bAaT++(wo $J_{FI}$)     &  & \checkmark                     & \checkmark                     & \textbf{11.15 (3.13$\%\uparrow$)}          &  & \checkmark & \checkmark & \textbf{6.30 (1.25$\%\uparrow$)}          &  & \checkmark & \checkmark & \textbf{7.73 (3.25$\%\uparrow$)} \\ \hline
			\end{tabular}%
		}
		\vspace{-5pt}
		\begin{flushleft}
			\footnotesize{ Notably, bAaT is enhanced with the first two constraints, but $J_{FI}$ to avoid disruption caused by Lipschitz discontinuity in Transformers\cite{kim2021lipschitz}.} 
		\end{flushleft}
		\vspace{-25pt}
	\end{table*}
	When solely utilizing $J_{MI}$ for alignment, it consistently outperforms previous versions that simply include $J_{Sim}$. This improvement is evident in models like DNN$\star$ and CNNLoc$\star$, which achieve significant reductions of 2.16 meters and 2.09 meters in MAE compared to their previous versions, \ie, DNN+ and CNNLoc+ in UJIIndoorLoc and Tampere, respectively. 
	Combined with $J_{Sim}$, it conspicuously enhances the MAE for all SoTA models, furnishing additional empirical support to fortify the alignment of acquired representations. For instance, simpler models like DNN++ and CNNLoc++ manifest more pronounced reductions in MAE across UJIIndoorLoc, UTS, and Tampere than in their previous versions. Even intricate models, such as Transformer-based architectures bAaT++ exhibit marginal yet discernible performance amelioration across all target datasets. We ascribe this enhancement to the adaptability to the relevance and quality of source datasets provided by the constraint, effectively capitalizing on cross-mutual information gleaned from source datasets. Furthermore, this also underscores the efficacy of the Expert Training phase, where neural generative models $G_i$, even of a simpler nature, can proficiently emulate representations acquired by all SoTA models, including sophisticated ones encapsulated by Transformers.

	\noindent$\triangleright$ \textbf{Effect of Functional Information.} 
	Unlike other constraints, $J_{FI}$ consistently leads to stable performance improvements for most SoTA networks. Interestingly, when combined with the remaining constraints, fully enhanced models significantly outperform the others, with lower errors. However, the inclusion of $J_{FI}$ can hinder the representation learning of certain models. In particular, bAaT's representation learning is negatively affected by $J_{FI}$ due to its non-compatibility with Lipschitz continuity caused by dot-product self-attention\cite{kim2021lipschitz}. Injecting such knowledge instead disrupts its functionality, especially the crucial self-attention mechanism, thus leading to undesired behavior. 
	
	The impact of the constraint on training data quantity is also discussed. Variations in dataset scale can affect the learning ability of models. In Tampere, the fully enhanced CNNLoc+++ model shows a 0.32 meters performance drop compared to CNNLoc++. This observation suggests that CNNLoc+++, which has been aligned with functional information from various surrogate teachers, may have become oversmoothed and unable to capture essential details in the Tampere dataset. In general, aligning such prior functional knowledge, namely, sparsity or smoothness from source datasets can deeply influence the model's behavior of adapting to the specific characteristics of the target dataset.

	\vspace{-5pt}
	\subsection{Impacts of Data Source Relevance}
	\vspace{-4pt}
	In this experimental scenario, we further investigate the impacts of the relevance of source datasets to the target dataset on the robustness of the proposed framework. For this purpose, various combinations of three benchmark datasets are all considered to create different levels of source relevance for verification. One of these datasets is selected as the target dataset, while the other two are used as source datasets alternately. 
	
	As observed in Tab.\ref{tab:util}, the experimental results show that the framework, most of the time, delivers reliable performance gains under varying relevance among source datasets to the target. Through all possible combinations, employing two source datasets simultaneously tends to bring out comparable or even better performance constantly.
	
	For example, there exist minor instances where convolutional models such as CNNLoc and BayesCNN exhibit slight performance deterioration when a single source dataset is used in UTS and UJIIndoorLoc, respectively. However, when both sources are employed, the performance deterioration is entirely mitigated, leading to significant performance gains, with notable improvements of up to 5.42$\%$, and 14.57$\%$ for CNN, and BayesCNN, respectively A better balance in alignment with a variety of reference points resulting from incorporating more source datasets may be linked to these stable benefits, which reduces the impact of specific environmental factor.
	Furthermore, the success can be attributed to the adaptability of the underlying cross-mutual and functional constraints $J_{MI}\&J_{FI}$, which are proportional to the relevance of source datasets to the target, thereby facilitating comprehensive alignment. These effects have been closely verified and reflected in Sect.\ref{subsec:abla}.

	\vspace{-8pt}
	\section{Conclusion}
	\vspace{-4pt}
	In this study, we contemplate an unexplored idea of elaborating a PnP knowledge-transfer framework that could be applicable straight to most SoTA localization models to consistently benefit from the comprehensive representation alignment between target RSS fingerprint datasets and public source datasets. 
	Diverging from conventional knowledge-transfer methods, this approach harnesses surrogate teachers to provide robust and environmentally insensitive representations for specialized models, while requiring neither access to source data nor modifications to existing architectures. Our extensive experiments demonstrate the framework's robustness to the source relevance and advocate the utilization of more independent source datasets to promote comprehensive alignment.
	\vspace{-8pt}
	\section{Acknowledgment}
	\vspace{-4pt}
	This publication is part of the project MOSAIC: enhancement of MicrOfluidic Sensing with deep symbolic Artificial IntelligenCe with file number 19985 of the research programme Open Technology Programme which is (partly) financed by the Dutch Research Council (NWO).
	
	This work made use of the Dutch national e-infrastructure with the support of the SURF Cooperative using grant no. EINF-6216.
	
	{\small
		\bibliographystyle{ieee_fullname}
		\bibliography{egbib}
	}
	
	\newpage
	
	\appendix
	\appendixpage 

	\section{Hyperparemeter Selection}
	For the angular margin $\alpha$, we searched within a range from 0 degrees (0 radians) to 30 degrees (0.52 radians) on UIJIndoorLoc, with a step size of 0.1 radians, to determine the optimal angular margin. Our experimentation revealed that an angular margin of 0.2 radians consistently yielded stable results. Higher values of the angular margin $\alpha$ 
	were not recommended, as they led to significant increases in model loss during training, making convergence extremely difficult.

	The weighting factors mentioned are also determined through a grid search on UIJIndoorLoc and subsequently applied to other datasets. This process is expedited by leveraging prior knowledge that the primary task of indoor localization (denoted as $J_{MAE}$) should be accorded greater emphasis, necessitating a larger range and higher amplitude for the weighting factor $\lambda_{1}$. Conversely, the high sensitivity of $J_{FI}$ to the learning ability necessitates a much narrower range for the weighting factor $\lambda_{4}$. To ensure balanced impacts, these weighting factors are normalized together. 
	
	The rationale behind conducting grid-search hyperparameter selection exclusively on the UJIIndoorLoc dataset before applying the chosen hyperparameters to other datasets lies in UJIIndoorLoc's notable generalizability. This dataset has a collection period spanning months and encompasses expansive campus coverage across three buildings, totaling 110,000 square meters, effectively capturing the dynamic nature of real-world environments. During the grid search process, we set specific search ranges for each hyperparameter, such as [1-5] for $\lambda_{1}$
	with a step size of 0.2, and [0.1-1] for $\lambda_{2,3,4}$
	with a step size of 0.1. Through this rigorous exploration of over 7 days, we identified that the set of values [3,0.5,0.5,0.5] yielded the best performance during testing on the UJIIndoorLoc dataset.
	
	\section{Incompatibly with other knowledge-transfer in RSS-Fingerprint-based Indoor 
		Localization}
	
	Compounded by the nature of radio propagation and multipath effects, the distinctive characteristics of RSS datasets, including variabilities in building structure, occupancy levels, and the arrangement and number of input WiFi anchors, give rise to uncompromising discrepancies in both appearance (input size) and content (locations). Unlike other data types such as images or text that easily achieve a common input size with minimal content alteration using standard interpolation techniques, RSS fingerprints cannot be resized as their arrangements of the disparate number of anchors are just unknown. Even if the input sizes were reluctantly synchronized, the content representing specific locations would undergo significant alterations, leading to deviations from the true data distribution. Consequently, traditional domain adaptation approaches\cite{farahani2021brief,pan2009survey}, such as meta-learning and adversarial learning, face limitations in their applicability to such datasets.
	
	Meta-learning approaches optimize a common meta-learner for the target task through the learning abilities of its versions trained on sub-tasks. However, implementing this approach to achieve a unified meta-learner for different RSS fingerprint datasets presents challenges. These datasets often vary significantly in the number of anchors, with differences of hundreds observed between datasets. Additionally, standard resizing techniques cannot be applied due to the unique characteristics of RSS fingerprints.
	
	Adversarial domain adaptation offers a potential solution to address differences in input size by employing separate feature generators for different datasets. However, this approach requires substantial modifications to existing architectures to ensure the delivery of homogeneous-sized features to domain discriminators for evaluation. Moreover, blindly learning domain-invariant features solely through artificial coarse-grained domain labels, especially in an adversarial learning framework, is inadequate for capturing fine-grained information particularly essential for precise localization. Additionally, this method is susceptible to instability, including notorious issues of model collapse, where discriminators fail to keep track of distribution changes in generated data.
	\section{Impacts of Target Relevance}
	
	\begin{table*}[th!]
		\caption{
			The impact of the target relevance to the source datasets on Expert Distilling phase}
		\label{tab:effi}
		\vspace{-8pt}
		\centering
		\resizebox{1.5\columnwidth}{!}{%
			\begin{tabular}{llcclclc}
				\hline
				\multicolumn{1}{c}{\multirow{3}{*}{\textbf{Method}}} &  & \multicolumn{2}{c}{\textbf{UJIIndoorLoc}} &  & \textbf{UTS}        &  & \textbf{Tampere}     \\ \cline{3-4} \cline{6-6} \cline{8-8} 
				\multicolumn{1}{c}{}                                 &  & 1\% (199/19937)     & 10\% (1993/19937)   &  & 10\% (910/9108)     &  & 10\% (69/697 )       \\ \cline{3-4} \cline{6-6} \cline{8-8} 
				\multicolumn{1}{c}{}                                 &  & MAE (m)$\downarrow$ & MAE (m)$\downarrow$ &  & MAE (m)$\downarrow$ &  & MAE (m) $\downarrow$ \\ \hline
				DNN\cite{felix2016fingerprinting}        &  & 160.56$\pm$0.24         & 18.28$\pm$0.41          &  & 10.65$\pm$5.52         &  & 25.04$\pm$14.99          \\
				DNN+++      &  & \textbf{36.36$\pm$2.32} & \textbf{17.42$\pm$0.45} &  & \textbf{8.18$\pm$0.68} &  & \textbf{24.28$\pm$10.87} \\ \hline
				CNNLoc\cite{song2019novel}     &  & 22.14$\pm$1.09 & 14.07$\pm$0.27          &  & 7.34$\pm$0.16 &  & 15.99$\pm$9.58          \\
				CNNLoc+++   &  & \textbf{21.93$\pm$1.17}          & \textbf{13.73$\pm$0.36} &  & \textbf{6.96$\pm$0.34}          &  & \textbf{12.78$\pm$1.09} \\ \hline
				BayesCNN\cite{sinha2021completely}   &  & 26.84$\pm$2.21          & 16.25$\pm$0.47 &  & 8.53$\pm$0.55          &  & 15.14$\pm$1.25          \\
				BayesCNN+++ &  & \textbf{25.3$\pm$1.93} & \textbf{15.68$\pm$0.72}          &  & \textbf{8.05$\pm$0.46} &  & \textbf{15.10$\pm$0.9} \\ \hline
				bAaT\cite{nguyen2023learning}       &  & 19.56$\pm$1.09          & 13.06$\pm$0.26          &  & 6.79$\pm$0.22          &  & 13.30$\pm$1.03          \\
				bAaT++ (wo $J_{FI}$)                                 &  & \textbf{18.18$\pm$0.86}      & \textbf{12.88$\pm$0.18}      &  & \textbf{6.58$\pm$0.18}       &  & \textbf{12.33$\pm$0.89}       \\ \hline
			\end{tabular}%
		}
	\end{table*}
	
	The robustness of the framework is additionally evaluated on the target side where the target distribution is changed with the proportion of training data. Specifically, experiments are first carried out using only $1\%$ and $10\%$ of the training data for UJIIndoorLoc, and then expanded to $10\%$ for the other datasets for general examination. This random partitioning is designed to simulate real-world scenarios for which all the models are subjected to the same conditions, and repeated for ten rounds to achieve statistical results. As demonstrated in Table \ref{tab:effi}, the framework exhibits its tolerance to target constrictions and consistently empowers state-of-the-art models to achieve strong performance.
	\section{Specific steps to the final $J_{MI}$ in Eq.4}
	\label{subsec:MI_explained}
	
	Cross-Mutual Information Maximization Constraint $J_{MI}$ in Eq.4 can be represented by JS Divergence $D_{JS}$ in retrospect as follows.
	\begin{equation}
		\scalebox{0.9}{
			$
			\begin{array}{l}
				{D_{JS}} = {\mathbb{E}_{Z_{{G_i}}^E,Z_S^E \sim p\left( {Z_{{G_i}}^E,Z_S^E} \right)}}\left( {\log\frac{{{p_{Z_{{G_i}}^EZ_S^E}}}}{{m\left( {Z_{{G_i}}^E,Z_S^E} \right)}}} \right) \\
				+ {\mathbb{E}_{Z_{{G_i}}^E \sim p\left( {Z_{{G_i}}^E} \right),Z_S^E \sim p\left( {Z_S^E} \right)}}\left( {\log\frac{{{p_{Z_{{G_i}}^E}}{p_{Z_S^E}}}}{{m\left( {Z_{{G_i}}^E,Z_S^E} \right)}}} \right)\\
				\\
				{\rm{where\,\,}}m\left( {Z_{{G_i}}^E,Z_S^E} \right) = \frac{1}{2}p\left( {Z_{{G_i}}^E} \right)p\left( {Z_S^E} \right) + \frac{1}{2}p\left( {Z_{{G_i}}^E,Z_S^E} \right)\\
				\\
				\Rightarrow 2{D_{JS}} = \\
				{\mathbb{E}_{Z_{{G_i}}^E \sim p\left( {Z_S^E} \right)}}\left[ \begin{array}{l}
					{\mathbb{E}_{Z_{{G_i}}^E \sim p\left( {Z_{{G_i}}^E\left| {Z_S^E} \right.} \right)}}\left[ \begin{array}{l}
						\log \frac{{p\left( {Z_{{G_i}}^E\left| {Z_S^E} \right.} \right)}}{{p\left( {Z_{{G_i}}^E} \right)}}\\
						- \log \left( {1 + \frac{{p\left( {Z_{{G_i}}^E\left| {Z_S^E} \right.} \right)}}{{p\left( {Z_{{G_i}}^E} \right)}}} \right)
					\end{array} \right]\\
					+ {\mathbb{E}_{Z_{{G_i}}^E \sim p\left( {Z_{{G_i}}^E} \right)}}\left[ { - \log \left( {1 + \frac{{p\left( {Z_{{G_i}}^E\left| {Z_S^E} \right.} \right)}}{{p\left( {Z_{{G_i}}^E} \right)}}} \right)} \right]
				\end{array} \right] 
			\end{array}
			$
		}
		\label{eq:JSD_explained}
	\end{equation}
	In addition, we make use of the mutual information estimator ${\Psi _\theta }$ \cite{hjelm2018learning} to estimate the logarithm ratio between ${p\left( {Z_{{G_i}}^E\left| {Z_S^E} \right.} \right)}$ and ${p\left( {Z_{{G_i}}^E} \right)}$, represented by ${\Psi _\theta }\left( {Z_{{G_i}}^E,Z_S^E} \right) = \log \frac{{p\left( {Z_{{G_i}}^E\left| {Z_S^E} \right.} \right)}}{{p\left( {Z_{{G_i}}^E} \right)}}$. In the reverse direction, it can be interpreted as $ \frac{{p\left( {Z_{{G_i}}^E\left| {Z_S^E} \right.} \right)}}{{p\left( {Z_{{G_i}}^E} \right)}} = {e^{{\Psi _\theta }\left( {Z_{{G_i}}^E,Z_S^E} \right)}}$. Put it all together in Eq.\ref{eq:JSD_explained}, the JS Divergence is elaborated further, which is exactly the Mutual Information constraint $J_{MI}$ presented in this work:
	
	\begin{equation}
		\begin{array}{l}
			{D_{JS}} = {\mathbb{E}_{p\left( {Z_{{G_i}}^E,Z_S^E} \right)}}\left[ {\log \frac{{{e^{{\Psi _\theta }\left( {Z_{{G_i}}^E,Z_S^E} \right)}}}}{{1 + {e^{{\Psi _\theta }\left( {Z_{{G_i}}^E,Z_S^E} \right)}}}}} \right] \\
			- {\mathbb{E}_{p\left( {Z_{{G_i}}^E} \right)p\left( {Z_S^E} \right)}}\left[ {\log \left( {1 + {e^{{\Psi _\theta }\left( {Z_{{G_i}}^E,Z_S^E} \right)}}} \right)} \right]\\
			= {\mathbb{E}_{p\left( {Z_{{G_i}}^E,Z_S^E} \right)}}\left[ {\log \frac{1}{{{e^{ - {\Psi _\theta }\left( {Z_{{G_i}}^E,Z_S^E} \right)}} + 1}}} \right] \\
			- {\mathbb{E}_{p\left( {Z_{{G_i}}^E} \right)p\left( {Z_S^E} \right)}}\left[ {\log \left( {1 + {e^{{\Psi _\theta }\left( {Z_{{G_i}}^E,Z_S^E} \right)}}} \right)} \right]\\
			= {\mathbb{E}_{p\left( {Z_{{G_i}}^E,Z_S^E} \right)}}\left[ { - \log \left( {1 + {e^{ - {\Psi _\theta }\left( {Z_{{G_i}}^E,Z_S^E} \right)}}} \right)} \right] \\
			- {\mathbb{E}_{p\left( {Z_{{G_i}}^E} \right)p\left( {Z_S^E} \right)}}\left[ {\log \left( {1 + {e^{{\Psi _\theta }\left( {Z_{{G_i}}^E,Z_S^E} \right)}}} \right)} \right]
		\end{array}
	\end{equation}

	\section{Specific steps expanded in Eq.9}
	We elaborate on the expression of the transmitting matrix $T_j$, which is mentioned in Eq.8 and is reduced to a simple form in Eq.9 as follows:
	\begin{equation}
		\begin{array}{l}
			{T_i} \buildrel \Delta \over = {\left[ {{{\left( {\widehat {Z}_{j - 1}} \right)}^\Gamma }\left( {\widehat {Z}_j} \right)} \right]^\Gamma }\left[ {{{\left( {\widehat {Z}_{j - 1}} \right)}^\Gamma }\left( {\widehat {Z}_j} \right)} \right]\\
			= {\left( {{W_j}\widehat {Z}_{j - 1}} \right)^\Gamma }\left( {\widehat {Z}_{j - 1}} \right){\left( {\widehat {Z}_{j - 1}} \right)^\Gamma }\left( {{W_i}\widehat {Z}_{j - 1}} \right)\\
			= {\left( {{W_j}\widehat {Z}_{j - 1}} \right)^\Gamma }\left( {{W_j}\widehat {Z}_{j - 1}} \right)\\
			= {\widehat {Z}_{j - 1}^\Gamma }\left( {{W_j}^\Gamma {W_j}} \right)\widehat {Z}_{j - 1}
		\end{array}
	\end{equation}
	
	\section{Power Iteration Algorithm}
	The proposed framework estimates the spectral norm of the blocks in neural networks using the Power Iteration Algorithm. This method is chosen for its lightweight computation and continuous differentiability. Here is the pseudocode for the algorithm:
	\begin{algorithm}[!ht]
		\DontPrintSemicolon
		
		\KwInput{Transmitting Matrix $T_i$, power iteration $n$}
		\KwOutput{The Largest Eigenvalue $\sigma(\cdot)$ } 
		v = tf.random.normal([$T_i$.shape[0], $T_i$.shape[1]]) \;
		\For{ $i=0$ $\to$ $n$}
		{
			$m$ $=$ tf.matmul($T_i$, $v$) \\
			
			$\mu$ $=$ tf.sqrt( tf.reduce$\_$sum(tf.square($m$), axis $=1$))\\
			$v$ $=$ $m/ \mu$
		}
		$v\_norm =$ tf.sqrt(tf.reduce$\_$sum(tf.square($v$), axis $= 1$))\\
		$\sigma(T_i) =$ tf.sqrt($\mu / v\_norm$)\\
		\Return $\sigma(T_i)$
		\caption{Numerical Estimation of Spectral Norm  with Tensorflow Pseudocode, called $top\_eigenvalue$}
	\end{algorithm}
	
	\section{ Pseudo code of the training pipeline}
	The framework executes the alignment through two primary phases. The first phase called \textit{Expert Training}, involves modeling the representations established by the specialized networks on their respective source datasets using surrogate teacher networks. In the second phase, \textit{Expert Distilling}, these modeled representations are collectively distilled into essential knowledge for alignment with representations learned on the target dataset.
	\begin{algorithm}[!ht]
		\DontPrintSemicolon
		
		\KwInput{Surrogate teachers $\{{G_i}\}_{i=1}^{N}={\{F_i^G,E_i^G\}}_{i=1}^N$,\\ Critics $\{C_i\}_{i=1}^N$, $c\_step$, Source dataset ${D}_{i=1}^N$,\\ Specialized models $\{{S_i}\}_{i=1}^N={\{F_i^S, E_i^S,R_i^S\}}_{i=1}^N$, \\ Epoch $E$, gradient weight $\alpha$, loss weights $\{\beta_i\}_{i=1}^3$\\}
		\KwOutput{ Pre-trained surrogate teachers $\{G_i\}_{i=1}^N$ } 
		Initialize $T\_list$ \;
		\For{ $i=1$ $\to$ $N$}
		{   \For{ epoch $e=1$ $\to$ $E$}
			{   
				Load $D_i$\\
				Initialize  $G_i$, $C_i$, $S_i$\\
				\For{ batch $b$ $\in$ $D_i$}
				{   \tcc{First, training Critics for $c$ steps}
					\For{$k$ to $c\_step$}
					{
						Initialize $Noise$\\
						$Z^F_{S_i}, Z^E_{S_i}, {\hat {Y}_{S_i}} = S_i(b)$\\
						$Z^F_{G_i}, Z^E_{G_i}=G_i(Noise)$\\
						$r\_logits = C_i(Z^E_{S_i})$\\
						$f\_logits = C_i(Z^E_{G_i})$\\
						$gp=grad\_penalty(C_i, b, Z^E_{S_i}, Z^E_{G_i})$\\
						$L_c = critic\_loss(r\_logits,f\_logits)$\\
						$L_{tc}=L_c+\alpha*gp$\\
						$C_i = update(L_{tc},C_i)$\\
						
					}
					
					\tcc{Then, training Generators and Specialized model}
					Initialize $Noise$\\
					
					$Z^F_{S_i}, Z^E_{S_i}, {\hat {Y}_{S_i}} = S_i(b)$\\
					
					$Z^F_{G_i}, Z^E_{G_i}=G_i(Noise)$\\
					
					${\hat {Y}_{G_i}} = R^S_i(Z^E_{G_i})$\\
					
					$L_{S\_MAE}=J_{MAE}({\hat {Y}_{S_i}}, Y_{S_i})$\\
					
					$L_{G\_MAE}=J_{MAE}({\hat {Y}_{G_i}}, Y_{S_i})$
					
					$L_{Sim}=J_{Sim}(Z^E_{S_i}, Z^E_{G_i})$\\
					
					$L_{tg}=\beta_1*L_{S\_MAE}+\beta_2*L_{G\_MAE}+\beta_3*L_{Sim}$
					
					$G_i = update(L_{tg},G_i)$\\
					
					$S_i = update(L_{S\_MAE},S_i)$\\
				}
			}
			$T\_list.append(G_i)$
		}
		
		\Return $T\_list$
		\caption{Expert Training Phase}
	\end{algorithm}
	\begin{algorithm}
		\DontPrintSemicolon
		\KwInput{Surrogate teachers $\{{G_i}\}_{i=1}^N={\{F_i^G,E_i^G\}}_{i=1}^N$,\\ Target dataset ${D}_t$ \\Mutual Information Estimator $\Psi_\theta$,\\ Specialized models $S={\{F^S, E^S,R^S\}}$, \\  Epoch $E$, loss weights $\lambda_{i=1}^4$\\}
		\KwOutput{Specialized models $S={\{F^S, E^S,R^S\}}$ }
		Load $D_t$\\
		Initialize ${\Psi _{\theta}}$, $S$\\
		\For{ $e=1$ $\to$ $E$}
		{
			\For{ $b$ $\in$ $D_t$}
			{
				Initialize $Noise$, $L^{MI}_t$, $L^{Sim}_t$, $L^{FI}_t$\\
				
				$Z^F_{S}, Z^E_{S}, {\hat {Y}_{S}} = S(b)$\\
				
				$L_{S\_MAE}=J_{MAE}({\hat {Y}_{S_i}}, Y_{S_i})$\\
				
				\tcc{Computing Functional Information in the specialized model \textbf{S} for comparison with other surrogate teachors}
				$TM_S=transmitting\_matrix(Z^F_{S},Z^E_{S})$\\
				$FI_S=top\_eigenvalue(TM_S)$
				
				\For{ $i=1$ $\to$ $N$}
				{
					Load $G_i$\\
					$Z^F_{G_i}, Z^E_{G_i}=G_i(Noise)$\\
					\tcc{Computing Mutual Information Constraint}
					$product\_examples=concat(Z^E_{S_i}[1:],Z^E_{S_i}[0])$\\
					$joint\_stat={\Psi _{\theta}}(Z^E_{S_i},Z^E_{G_i})$\\
					$product\_stat={\Psi _{ \theta}}(product\_examples,Z^E_{G_i})$\\
					$L^{MI}_i=J_{MI}(joint\_stat,product\_stat)$\\
					$L^{MI}_t+=L^{MI}_i$\\\tcc{Computing Angular Similarity Constraint}
					
					$L^{Sim}_i=J_{Sim}(Z^E_{S_i},Z^E_{G_i})$\\
					$L^{Sim}_t+=L^{Sim}_i$\\
					
					\tcc{Computing Functional Information Constraint}
					$TM_{G_i}=transmitting\_matrix(Z^F_{G_i},Z^E_{G_i})$\\
					$FI_{G_i}=top\_eigenvalue(TM_{G_i})$\\
					
					$L^{FI}_i=J_{FI}(FI_S,FI_{G_i})$\\
					$L^{FI}_t+=L^{FI}_i$\\
					
				}
				
				$L_{t} = J_{overall}(\lambda_{i=1}^4,L_{S\_MAE},L^{Sim}_t,L^{MI}_t,L^{FI}_t)$\\
				$S=update(L_{t},S)$
				
			}
		}
		\Return $S={\{F^S, E^S,R^S\}}$
		\caption{Expert Distilling Phase}
	\end{algorithm}
	
\end{document}